\documentclass[
dvipsnames, table,   %
format=acmsmall,     %
anonymous=false,     %
authorversion=true, %
]{acmart}

\usepackage{algcompatible, algorithm}
\usepackage[noend]{algpseudocode}

\usepackage{xspace}

\usepackage[]{subcaption}
\DeclareCaptionSubType*{algorithm}

\usepackage{tabularx, siunitx}

\usepackage{tikz}
\usetikzlibrary{arrows, decorations.markings}

\newcommand\eFront{$\epsilon$-PF\xspace}
\newcommand\eRandom{$\epsilon$-RS\xspace}

\newcommand{\best}{{\cellcolor[gray]{0.75}}}
\newcommand{\statsimilar}{{\cellcolor[gray]{0.9}}}

\newcommand\mX{\mathcal{X}}
\newcommand\Real{\mathbb{R}}

\newcommand\mP{\mathcal{P}}
\newcommand\Papprox{\tilde{\mathcal{P}}}

\newcommand\mGP{\ensuremath{\mathcal{GP}}\xspace}
\newcommand\mf{\mathbf{f}}
\newcommand\mD{\mathcal{D}}
\newcommand\mN{\mathcal{N}}

\newcommand\mS{\mathcal{S}}
\newcommand\EI{\alpha_{EI}}
\newcommand\WEI{\alpha_{WEI}}

\newcommand\PI{\alpha_{PI}}
\newcommand\UCB{\alpha_{UCB}}

\newcommand\prob{p}

\newcommand\expc{\mathbb{E}}

\DeclareMathOperator*{\argmax}{\arg\!\max}

\DeclareMathOperator{\LatinHypercubeSampling}{LatinHypercubeSampling}

\newcommand{\trp}{^\top}
\newcommand{\given}{\,|\,}
\newcommand{\bx}{\mathbf{x}}

\newcommand{\bff}{\mathbf{f}}

\newcommand{\fstar}{f^\star}
\newcommand{\xnext}{\bx'}

\newcommand{\bkappa}{\boldsymbol{\kappa}}

\newcommand*{\eg}{e.g.\@\xspace}
\newcommand*{\ie}{i.e.\@\xspace}

\setcopyright{acmlicensed}
\copyrightyear{2021}
\acmYear{2021}
\acmDOI{10.1145/3425501}

\acmJournal{TELO}
\acmVolume{1}
\acmNumber{1}
\acmArticle{1}
\acmMonth{4}

\begin{document}

\title{Greed is Good: Exploration and Exploitation Trade-offs in Bayesian Optimisation}
      
\author{George {De Ath}}
\email{g.de.ath@exeter.ac.uk}
\orcid{0000-0003-4909-0257}
\affiliation{%
  \department{Department of Computer Science}
  \institution{University of Exeter}
  \city{Exeter}
  \country{United Kingdom}
}

\author{Richard M. Everson}
\email{r.m.everson@exeter.ac.uk}
\orcid{0000-0002-3964-1150}
\affiliation{%
  \department{Department of Computer Science}
  \institution{University of Exeter}
  \city{Exeter}
  \country{United Kingdom}
}

\author{Alma A. M. Rahat}
\email{a.a.m.rahat@swansea.ac.uk}
\orcid{0000-0002-5023-1371}
\affiliation{%
  \department{Department of Computer Science}
  \institution{Swansea University}
  \city{Swansea}
  \country{United Kingdom}
}

\author{Jonathan E. Fieldsend}
\email{j.e.fieldsend@exeter.ac.uk}
\orcid{0000-0002-0683-2583}
\affiliation{%
  \department{Department of Computer Science}
  \institution{University of Exeter}
  \city{Exeter}
  \country{United Kingdom}
}

\begin{abstract}
  The performance of acquisition functions for Bayesian optimisation to locate
  the global optimum of continuous functions is
  investigated in terms of the Pareto front between exploration and
  exploitation. We show that Expected Improvement (EI) and the Upper Confidence
  Bound (UCB) always select solutions to be expensively evaluated on the
  Pareto front, but Probability of Improvement is not guaranteed to do so
  and Weighted Expected Improvement does so only for a restricted range of
  weights.
  
  We introduce two novel $\epsilon$-greedy acquisition functions.
  Extensive empirical evaluation of these together with random search, purely
  exploratory, and purely exploitative search on 10 benchmark problems in 1 to 10
  dimensions shows that $\epsilon$-greedy
  algorithms are generally at least as effective as conventional
  acquisition functions (\eg EI and UCB), particularly with a limited budget. In higher
  dimensions $\epsilon$-greedy approaches are shown to have improved
  performance over conventional approaches. These results
  are borne out on a real world computational fluid dynamics optimisation
  problem and a robotics active learning problem. Our analysis and experiments
  suggest that the most effective strategy, particularly in higher dimensions,
  is to be mostly greedy, occasionally selecting a random exploratory solution.
\end{abstract}


          
\maketitle

\section{Introduction}

Global function optimisers search for the minimum or maximum of a function
by querying its value at selected locations. All optimisers must therefore
balance exploiting knowledge of the function gained from the evaluations
thus far with exploring other regions in which the landscape is unknown and
might hold better solutions. This balance is particularly acute when a
limited budget of function evaluations is available, as is often the case
in practical problems, \eg \citep{jones:ego, shahriari:ego}. Bayesian optimisation
is an effective  form of surrogate-assisted optimisation in which a probabilistic model
of the function is constructed from the evaluations made so far. The
location at which the function is next (expensively) evaluated is chosen as
the location which maximises an \textit{acquisition function} which makes the
balance between exploration and exploitation explicit by combining the predicted
function value at a location with the uncertainty in that prediction.

Here we regard the balance between exploration and exploitation as
itself a two-objective optimisation problem. We show that many, but not
all, common acquisition functions effectively select from the Pareto front between objectives
quantifying exploration and exploitation. In common with \citep{bischl:mbo,
zilinskas:tradeoff, grobler:simple,feng:egomo}, we propose choosing the next location to be
expensively evaluated from the estimated Pareto set of solutions found by a
two-objective evolutionary optimisation of the exploration and exploitation
objectives. We compare the performance of various methods for selecting
from the estimated Pareto front and propose two new $\epsilon$-greedy schemes that usually
choose the solutions with the most promising (exploitative) value, but occasionally
use an alternative solution  selected at random from either the estimated Pareto set
or the entire feasible space. 

Our main contributions can be summarised as follows:
\begin{itemize}
\item We present a unified analysis of common acquisition functions in terms 
      of exploration and exploitation and give the first detailed analysis of
      weighted expected improvement.
\item We investigate the use of the exploration-exploitation trade-off front in
      selecting the next location to expensively evaluate in Bayesian 
      optimisation.
\item We present two novel $\epsilon$-greedy acquisition functions for Bayesian
      optimisation as well as other acquisition functions that use the 
      exploration-exploitation trade-off front.
\item These methods are empirically compared on a variety of synthetic test
      problems and two real-world applications.
\item We demonstrate that the $\epsilon$-greedy approaches are at least as
	  effective as the conventional acquisition functions on lower-dimensional
	  problems and become superior as the number of decision variables increases.
\end{itemize}

We begin in Section~\ref{sec:bayes-optim} by briefly reviewing Bayesian
optimisation together with  Gaussian processes --- which are commonly used for
surrogate modelling of the function. We pay particular attention to
acquisition functions  and the way in which they balance exploration and exploitation.
The exploration-exploitation trade-off is viewed through the lens of
multi-objective optimisation in Section~\ref{sec:expl-expl-trade}, which
leads to the proposed $\epsilon$-greedy schemes in 
Section~\ref{sec:epsil-greedy-appr}. Extensive empirical evaluations on well-known
test problems are presented in Section~\ref{sec:experiments}, along with comparisons on a real world computational fluid dynamics optimisation and robot active learning problem.

\section{Bayesian Optimisation}
\label{sec:bayes-optim}

\textit{Bayesian optimisation (BO)}
is a particular method of surrogate-assisted  optimisation.
In practice, it has proved to be a very effective approach for single
objective expensive optimisation problems with limited budget on the number
of true  function evaluations. It was first proposed by 
\citet{kushner:ego} in the early 1960s, and later improved and popularised
by \citet{mockus:ei} and \citet{jones:ego}. 
A recent review of the topic can be found in \citep{shahriari:ego}.

Without loss of generality, the optimisation problem may be expressed as:
\begin{align}
\max_{\bx \in \mX} f(\bx),
\end{align}
where  $\mX \subset \Real^d$
is the feasible space and $f:\Real^d \rightarrow \Real$. 
Algorithm~\ref{alg:bo} outlines the standard Bayesian optimisation procedure. In
essence, it is a global search strategy that sequentially samples the design
space at likely locations of the global optimum taking into account not
only the predictions of the surrogate model but also the 
uncertainty inherent in modelling the unknown function to be optimised
\citep{jones:ego}. It starts
(line~\ref{alg-bo-lhs}) with a space filling design (\eg Latin hypercube
sampling \citep{mckay:lhs}) of the parameter space, constructed independent
of the function space. The samples $ X = \{\bx_t\}_{t=1}^M$ from this
initial design are then (expensively) evaluated with the function,
$f_t = f(\bx_t)$, to construct a training dataset from which the surrogate
model may be learned. We denote the vector of evaluated samples by $\bff$.
Then, at each iteration of the main part of the algorithm, a regression
model is trained using the function evaluations obtained thus far 
(line~\ref{alg:train}). In Bayesian optimisation the regression model is used to
predict the most likely value of $f(\bx)$ at new locations, but also the
uncertainty in the model estimate. In common with most work on Bayesian
optimisation, we use Gaussian process models (GPs), which subsume Kriging models,
as regressors; these are described in Section~\ref{sec:modell-with-gauss}.
The choice of where to next evaluate $f$ is made by finding the location
that maximises 
an \textit{acquisition function} or \textit{infill} criterion $\alpha(\bx, \mD, \theta)$ which balances exploitation of
good regions of design space found thus far with the exploration of
promising regions indicated by the uncertainty in the surrogate's
prediction. Various common infill criteria are discussed and analysed from
a multi-objective point of view in Section~\ref{sec:infill-criteria}. The
design maximising the infill criterion, $\xnext$ is often found by an
evolutionary algorithm (line~\ref{alg:opt_infill}), which is able to
repeatedly evaluate the computationally cheap infill criterion. Finally,
$f(\xnext)$ is expensively evaluated and the training data $(X, \bff)$
augmented with $\xnext$ and $f(\xnext)$ (lines~\ref{alg:expensively-evaluate} 
to~\ref{alg:update-f}).
The process is repeated until the budget is exhausted.

\begin{algorithm} [t!]
  \caption{Standard Bayesian optimisation.}
  \label{alg:bo}

 \begin{algorithmic}[]
 	\State \textbf{Inputs:}
 	\State {\setlength{\tabcolsep}{2pt}%
 	        \begin{tabular}{c p{2pt} l}
 			$M$ &:& Number of initial samples \\
 			$T$ &:& Budget on the number of expensive evaluations
 			\end{tabular}
 	       }%
  \end{algorithmic}
  \medskip
 
  \begin{algorithmic}[1]
    \Statex \textbf{Steps:}
	\State $X \gets \LatinHypercubeSampling(\mX, M)$ \label{alg-bo-lhs}
        \Comment{\small{Generate initial samples}}
        \For{$t = 1 \rightarrow M$}
        \State $f_t \gets f(\bx_t)$ \Comment{\small{Expensively evaluate all initial samples}}
        \EndFor
	
	\State $\mD \gets \{(X, \bff) \}$
	\For{$t = M+1 \rightarrow T$}
        \State  $\theta \gets  \text{Train\mGP}(\mD)$ \label{alg:train} \Comment{\small{Train a \mGP model}} 
        \State \label{choose-xnext} $\xnext \gets \argmax_{\bx \in \mX}~
        \alpha (\bx, \mD, \theta)$\Comment{\small{Maximise infill
            criterion}\label{alg:opt_infill}}
        \State $f' \gets f(\bx')$ \label{alg:expensively-evaluate}
        \Comment{Expensively evaluate $\xnext$}
        \State $X \gets X \cup \{\xnext\}$ \label{alg:update-X} \Comment{Augment data}
        \State $\mf \gets \mf \cup
        \{f'\}$ \label{alg:update-f}
        \State $\mD \gets \{(X, \bff) \}$
	\EndFor
	\State \Return $\mD$
  \end{algorithmic}
\end{algorithm}

\subsection{Modelling with Gaussian Processes}
\label{sec:modell-with-gauss}

Gaussian processes are commonly used to construct a surrogate
model of $f(\bx)$ and we therefore briefly describe them here; a
comprehensive introduction may be found in \citep{rasmussen:gpml}.
In essence, a GP is a collection of random variables, and any finite
number of these have a joint Gaussian distribution \citep{rasmussen:gpml}.
With data comprising $f(\bx)$ evaluated at $M$ locations
$\mD = \{(\bx_m, f_m \triangleq f(\bx_m))\}_{m = 1}^M$, the
predictive probability for $f$ at $\bx$ is a Gaussian distribution with mean
$\mu(\bx)$ and variance $\sigma^2(\bx)$:
\begin{align}
\label{eq:gpmod}
 \prob (f \given  \bx, \mD, \theta) = \mN(\mu(\bx), \sigma^2(\bx)),
\end{align}
where the mean and variance are 
\begin{align}
     \mu(\bx) &= \bkappa(\bx, X) K^{-1} \mf \label{eq:gp-mu} \\
\sigma^2(\bx) & =\kappa(\bx, \bx) - \bkappa(\bx, X)\trp K^{-1} \kappa(X, \bx). \label{eq:gp-sigma}
\end{align}
Here $X \in \Real^{M \times d}$ is the matrix of design locations and
$\bff \in \Real^M$ is the corresponding vector of the true function
evaluations; thus $\mD = \{(X, \mf)\}$. The covariance matrix
$K\in \Real^{M\times M}$ represents the covariance function
$\kappa(\bx, \bx'; \theta)$ evaluated for each pair of observations and
$\bkappa(\bx, X) \in \Real^M$ is the vector of covariances between $\bx$
and each of the observations; $\theta$ denotes the kernel
hyperparameters. We use a flexible class of covariance
functions embodied in the Mat{\'e}rn $5/2$ kernel, as recommended for modelling
realistic functions \citep{snoek:practical}. 
Although it is  beneficial to marginalise $\theta$ with respect
to a prior distribution, here we follow standard practise and fix on a
single value of the hyperparameters by maximising the log likelihood each time
the data is augmented by a new expensive evaluation:\footnote{We use the L-BFGS
algorithm with $10$ restarts to estimate the hyper-parameters \citep{gpy}.}
\begin{equation}
\log \prob(\mD\given \theta) = - \frac{1}{2} \log |K| - \frac{1}{2} \mf^\top K^{-1} \mf - \frac{M}{2} \log (2\pi).
\end{equation}
Henceforth, we omit $\theta$ for notational simplicity, and assume that these 
are set by maximum  likelihood estimates.

\subsection{Infill Criteria and Multi-Objective Optimisation}
\label{sec:infill-criteria}

An infill criterion or acquisition function $\alpha(\bx, \mD, \theta)$ is a measure
of quality that enables us to decide which locations $\bx$ are promising
and consequently where to expensively evaluate $f$. It is based on the
prediction  $\prob (f \given  \bx, \mD)$ %
from the surrogate (GP) model that represents our
belief about the unknown function $f$ at a decision vector $\bx$ based on
the $M$ observations $\mD$. Although $\alpha(\bx, \mD, \theta)$ depends on $\mD$
and on the hyperparameters ($\theta$) of the GP, for clarity we
suppress this dependence and write $\alpha(\bx)$. The predictive
distribution~\eqref{eq:gpmod} is Gaussian, with mean and variance given 
by~\eqref{eq:gp-mu} and~\eqref{eq:gp-sigma}. The predicted mean and uncertainty
enable an infill criterion to strike a balance between myopic exploitation
(concentrating on regions where the mean prediction $\mu(\bx)$ is large) and global
exploration (concentrating on regions where the uncertainty $\sigma(\bx)$ about $f$ is
large). Since, in general both exploitation and exploration are desirable,
we may view these as competing criteria: a location $\bx$ that is both more
exploitative and more exploratory than an alternative $\bx'$ is to be
preferred over $\bx'$. Using the notation of multi-objective optimisation,
 a location $\bx$ dominates $\bx'$, written $\bx \succ \bx'$,
iff $\mu(\bx) \ge \mu(\bx')$ and $\sigma(\bx) \ge \sigma(\bx')$ and they are not equal on both. We present BO procedures that select solutions from the Pareto
optimal set of locations, namely those which are not dominated by any other feasible
locations:
\begin{align}
  \label{eq:pareto-set}
  \mathcal{P} = \{ \bx \in \mathcal{X} \given  \bx' \not\succ \bx 
  \, \forall \bx' \in \mathcal{X} \},
\end{align}
where $\bx' \not\succ \bx$ indicates that $\bx'$ does not dominate $\bx$.

\begin{figure}[t]
  \centering
  \includegraphics[width=\textwidth]{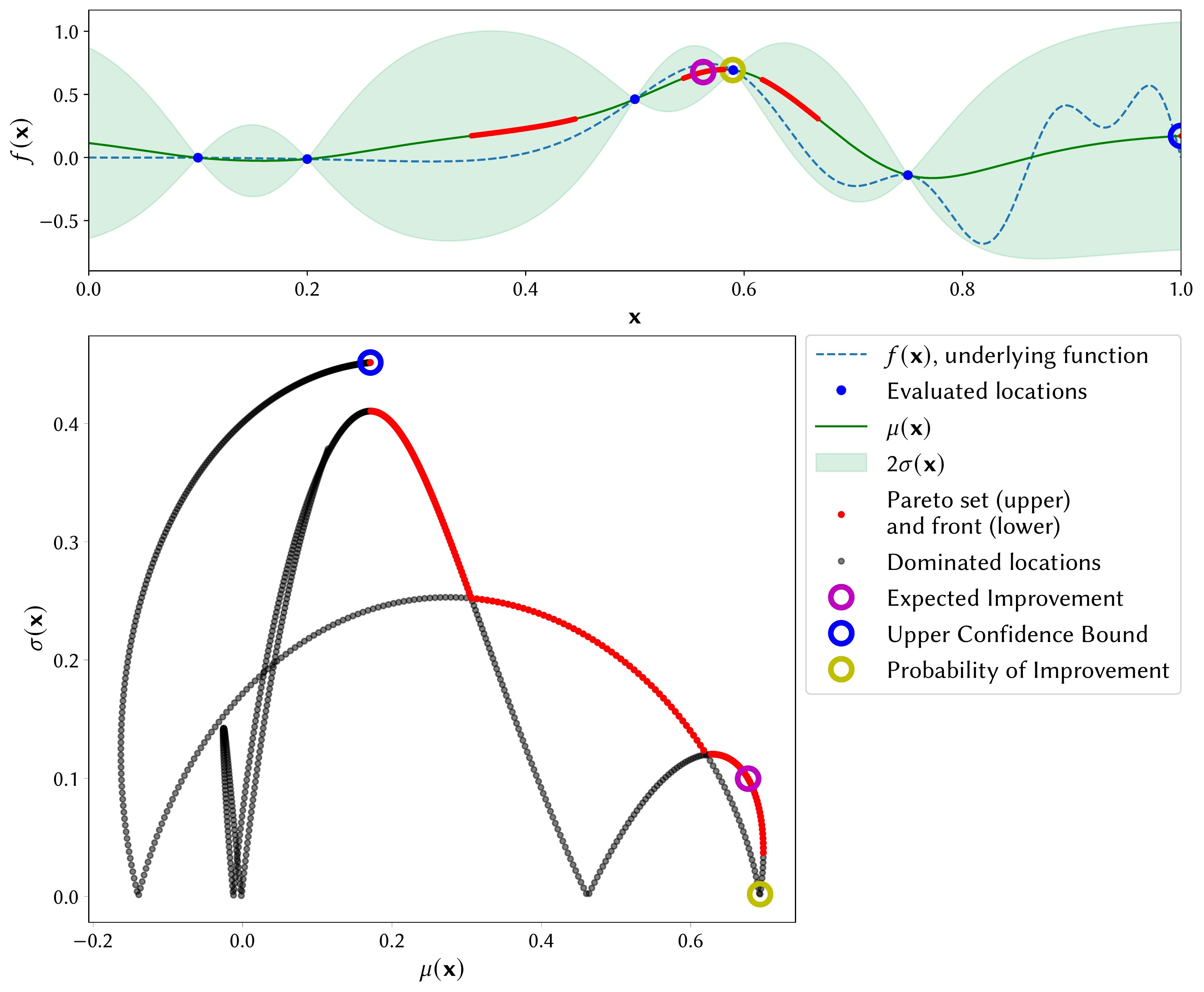}
  \caption{Example Pareto front: \textit{Top:} Gaussian Process
    approximation to a function (blue dashed curve) resulting from the 5
    observations shown; mean $\mu(x)$ is shown in dark green and twice the
    posterior standard deviation $\sigma(x)$ is shown as the light green
    envelopes. \textit{Bottom:} 200 samples uniformly spaced in
    $\mathcal{X}$ plotted in $\mu, \sigma$ space. The non-dominated
    locations forming the Pareto front are shown in red and their locations
    marked above. Locations maximising the Expected Improvement, Upper
    Confidence Bound and Probability of Improvement acquisition functions
    are marked in both plots.}
  \label{fig:PF-example}
\end{figure}

Figure~\ref{fig:PF-example} illustrates the approximate Pareto front, 
$\{ (\mu(\bx), \sigma(\bx)) \given \bx \in \mathcal{P} \}$, for a simple
one-dimensional function.  Note that the Pareto set is disjoint in $\mathcal{X}$
and in $(\mu, \sigma) $ space. The locations maximising three popular
acquisition functions, Expected Improvement (EI), Upper Confidence Bound (UCB)
and Probability of Improvement (PI) are highlighted. The maximisers of EI and 
UCB are elements of the Pareto set, whereas the maximiser of PI is not.

We now present some of the most popular acquisition functions  
used in BO, and discuss how they achieve a balance between exploration and
exploitation.

\subsubsection{Upper Confidence Bound.}
\label{sec:upper-conf-bound}

An optimistic policy, first proposed by \citet{lai:ucb}
is to overestimate the mean with added uncertainty: this is known as the
upper confidence bound infill criterion (UCB). A proof of convergence under
appropriate assumptions is given in \citep{srinivas:ucb}. The UCB acquisition function is  a weighted sum of the mean prediction and uncertainty:
\begin{align}
\UCB(\bx) = \mu(\bx) + \sqrt{\beta_t} \sigma (\bx), 
\end{align}
where $\sqrt{\beta_t} \ge 0$ is the weight, which generally depends upon
the number of function evaluations, $t$. The addition of a multiple of the uncertainty
means that the criterion prefers locations where the mean is large 
(exploitation) or mean combined with the uncertainty is sufficiently large
to warrant exploration.  

When $\beta_t = 0$ UCB becomes a purely exploitative scheme and therefore
the solution with the best predicted mean is evaluated expensively. Thus,
it may rapidly converge to a local maximum prematurely. In contrast, when
$\beta_t$ is large, the optimisation becomes purely exploratory, evaluating
the location where the posterior uncertainty (variance) is largest, which is equivalent
to maximally reducing the overall predictive entropy of the model \citep{srinivas:ucb}.
Consequently, it may eventually locate the global optima, but the rate of 
convergence may be very slow.

Some authors suggest tuning $\beta_t$ during the course of the optimisation
\citep{shahriari:ego}; indeed \citeauthor{srinivas:ucb}'s convergence proof
depends on a particular schedule in which $\sqrt{\beta_t}$ increases like the
logarithm of $t$, so that more weight is given to exploratory moves as the optimum is approached \citep{srinivas:ucb}.

Clearly, UCB increases monotonically as either the mean prediction $\mu$ or
the uncertainty $\sigma$ increase; see Figure~\ref{fig:infill-contour}.
Consequently, if a set $\mS$ of candidate locations for expensive
evaluation is available and $\UCB$ is used to select the location with
maximum upper confidence bound, $\xnext = \argmax_{\bx \in \mS} \UCB(\bx)$,
then $\xnext$ is a member of the maximal non-dominated subset of $\mS$;
that is, there is no element of $\mS$ that dominates $\xnext$. We note
however, that although UCB selects a non-dominated location, there will
generally be other non-dominated locations that trade-off exploration and
exploitation differently.

\subsubsection{Expected Improvement.}
\label{sec:expected-improvement}

\begin{figure}[t]
  \centering
  \includegraphics[width=\textwidth, clip, trim={11 5 5 5}]{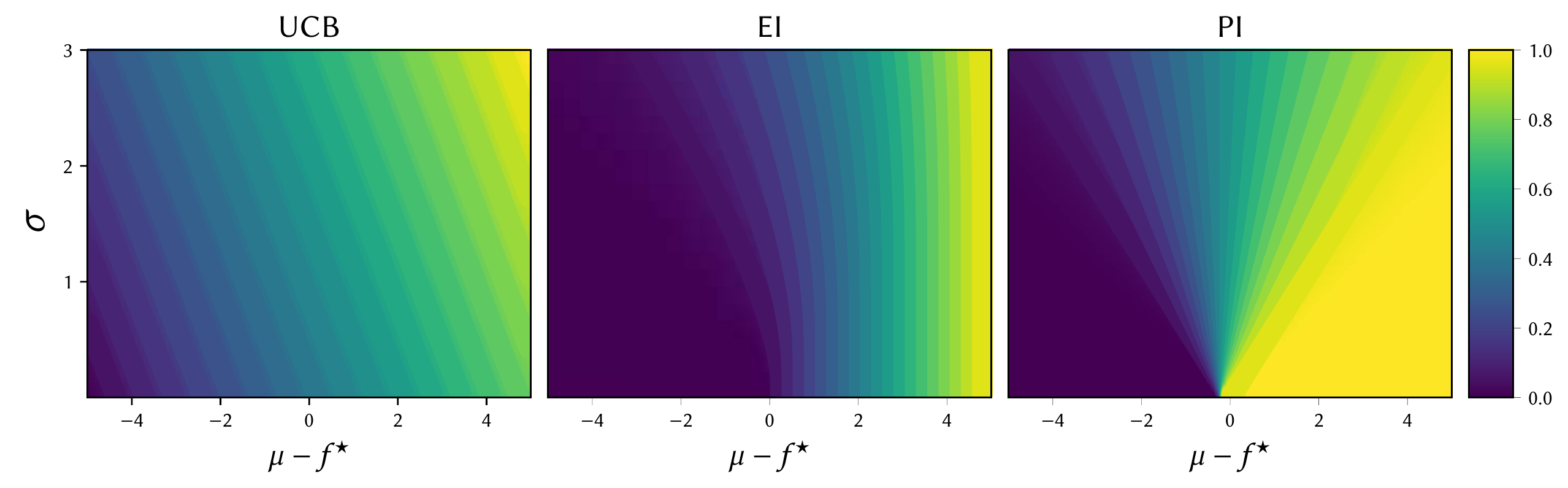}
  \caption{Contours of  upper confidence bound (UCB, $\beta_t=1$), expected
    improvement  (EI) and  probability of
    improvement (PI) as functions of
    predicted mean $\mu$ and uncertainty $\sigma$.   Since the scale of
    $\alpha$ is immaterial, all three infill criteria have been mapped to
    $[0, 1]$.}
  \label{fig:infill-contour}
\end{figure}

The expected improvement (EI) is perhaps the most popular infill criterion 
and is very widely used. It was first proposed by
\citet{mockus:ei}, and further developed by 
\citet{jones:ego}. \citeauthor{bull:convergence} has shown that, under certain
conditions, BO using EI is guaranteed to converge to the global optimum
\citep{bull:convergence}.

EI is based on the positive predicted improvement over the best solution
$\fstar = \max_m \{f_m \}$ observed so far.  If $\hat{f} = f(\bx)$ is an
evaluation of $f$ at $\bx$ then the improvement is 
\begin{align}
I(\bx, \hat{f}, \fstar)  = \max(\hat{f} - \fstar,~0).
\end{align}
Then the expected improvement at $\bx$  may be expressed as \citep{jones:ego}:
\begin{align}
  \label{eq:ei}
 \EI(\bx) = \expc [ I(\bx, \fstar)]
	&= \int_{-\infty}^{\infty} I(\bx, \hat{f}, \fstar) p(\hat{f} \given \bx, \mD)\,d\hat{f}  \nonumber \\
	&= \sigma(\bx) \left( s\Phi( s) + \phi (s)\right),
\end{align}
where $s = (\mu(\bx)-\fstar) / \sigma(\bx)$ is the predicted
improvement at $\bx$ normalised by the uncertainty, and $\phi(\cdot)$ and
$\Phi(\cdot)$ are the standard Gaussian probability density  and  cumulative
density functions. The infill criterion is therefore the improvement averaged
with respect to the posterior predictive probability of obtaining it.
Thus, EI balances the exploitation of solutions which are very likely to be
a little better than $\fstar$ with the exploration of others which may,
with lower probability, turn out to be much better.

As illustrated in Figure~\ref{fig:infill-contour}, $\EI(\bx)$ is monotonic
with respect to increase in both exploration, $\sigma$, and exploitation,
$\mu$. This can be seen by noting that
\begin{align}
\frac{\partial \EI }{\partial \mu} = \Phi(s)\quad \text{and} \quad
  \frac{\partial \EI}{\partial \sigma} = \phi(s)
\end{align}
are both positive everywhere \citep{jones:ego}. Consequently, like UCB, if
the next location to be expensively evaluated is selected by maximising EI,
the location will belong to the Pareto set maximally trading-off
exploration and exploitation. 

\subsubsection{Weighted Expected Improvement.}
\label{sec:WEI}

Some authors \citep{sobester05design, feng:egomo}  have associated the term,
$ \sigma(\bx) s\Phi( s) = (\mu(\bx)-\fstar)\Phi(s)$, in~\eqref{eq:ei} with 
the exploitation inherent in adopting $\bx$ as the next place to
evaluate.  Similarly,
the term $\sigma(\bx) \phi (s)$  has been associated with the
exploratory component. 
To control the balance between exploration and exploitation 
\citet{sobester05design} define an acquisition function that weights
these two terms differently: 
\begin{align}
  \label{eq:ei-omega}
 \WEI(\bx, \omega) 
	&= \sigma(\bx) \left[ \omega s\Phi( s) + (1-\omega)\phi (s)\right],
\end{align}
where $0\le\omega\le 1$.

However, it turns out that if the next point
for expensive evaluation is selected by maximising $\WEI(\bx)$ in some set
$\mS$ of candidate solutions,
$\bx' = \argmax_{\bx \in \mS} \WEI(\bx, \omega)$, then this only results in  choosing $\bx'$
in the maximal non-dominated set of $\mS$ for a relatively small range of $\omega$. This may be seen by considering
the partial derivatives of $\WEI(\bx, \omega)$. Without loss of generality,
we take $\fstar = 0$, so that $s = \mu/\sigma$. Then
\begin{align}
  \label{eq:dWEIdsigma}
  \frac{\partial \WEI }{\partial \sigma}
  & = -\omega s^2 \phi(s) + (1-\omega)(\phi(s) - s\phi'(s))\\
  &= 
    \left[
    1-\omega + (1-2\omega)s^2
    \right]\phi(s),
\end{align}
where we have used the fact that $\phi'(s) = -s\phi(s)$. Consequently, when
$\omega \le \tfrac{1}{2}$ the gradient 
$\frac{\partial \WEI }{\partial \sigma} > 0$ for all $s$. However, if
$\omega > \tfrac{1}{2}$ so that $1-2\omega < 0$ there are always regions
where $s = \mu/\sigma$ is sufficiently large that
$\frac{\partial \WEI }{\partial \sigma} < 0$. In
this case, there are therefore regions of $(\mu, \sigma)$ space in which
decreasing $\sigma$ increases $\WEI$, so
$\argmax_{\bx \in \mS} \WEI(\bx, \omega)$ is not guaranteed to lie in the Pareto set.

\begin{figure}[t]
\includegraphics[width=\textwidth, clip, trim={10 5 5 5}]{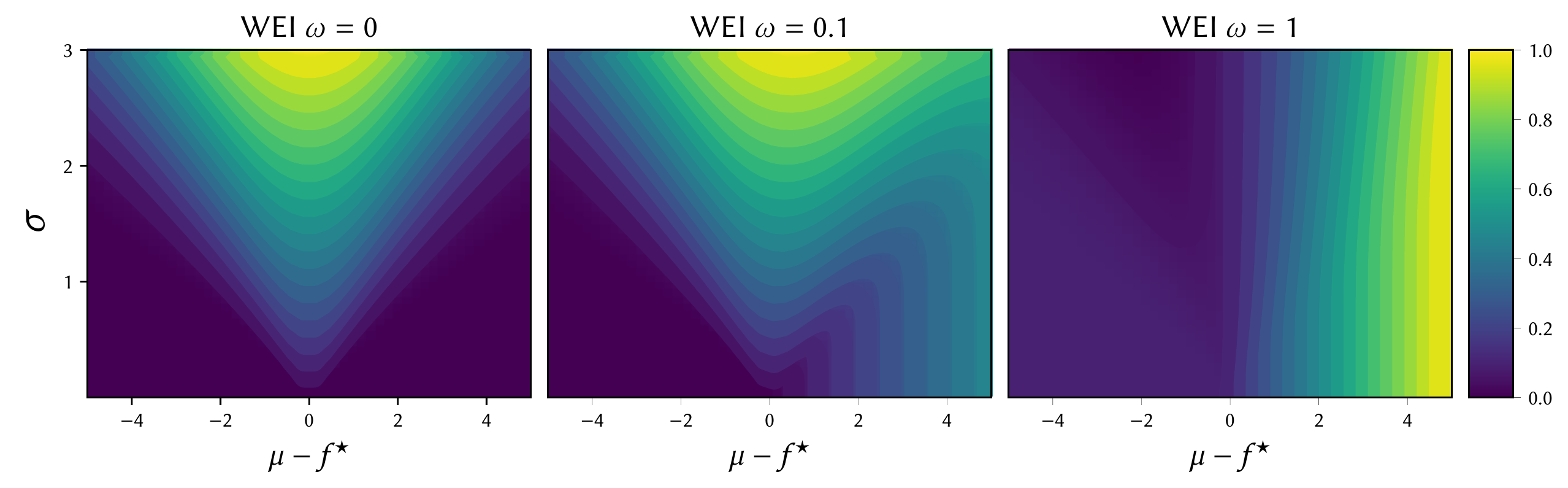}
\caption{Contours of weighted expected improvement as functions of the surrogate model's
predicted mean $\mu$ and uncertainty $\sigma$ for weights $\omega = 0, 0.1,
1$; equation~\eqref{eq:ei-omega}.  In none of these cases is the $\xnext$
maximising $\WEI(\xnext, \omega)$ guaranteed to lie in the Pareto set of
maximally exploratory and exploitative solutions.}
\label{fig:wei_contours}
\end{figure}

The gradient in the $\mu$ direction is
\begin{align}
  \label{eq:dWEIdmu}
  \frac{\partial \WEI }{\partial \mu} & = \omega \Phi(s) + (2\omega -1)s\phi(s).
\end{align}
Requiring that the gradient is non-negative, so that $\WEI$ is
non-decreasing with $\mu$ results in:
\begin{align}
  \label{eq:non-negative-gradient}
  \omega \ge (1-2\omega) \frac{s\phi(s)}{\Phi(s)}.
\end{align}
When $\omega > \frac{1}{2}$ it is straightforward to see 
that~\eqref{eq:non-negative-gradient} is always satisfied. The inequality is
also always satisfied for all $s < 0$ when $\omega < \tfrac{1}{2}$.  When
$\omega < \frac{1}{2}$ and $s \ge 0$ the inequality may be rewritten as
\begin{align}
  \label{eq:non-negative-rewritten}
  \frac{\omega}{1-2\omega} \ge \frac{s\phi(s)}{\Phi(s)}.
\end{align}
Defining
\begin{align}
  \label{eq:gamma-def}
  \gamma = \sup \frac{s\phi(s)}{\Phi(s)} \approx 0.295,
\end{align}
it can be seen that $\frac{\partial \WEI }{\partial \mu}$ is only
non-negative everywhere if $\omega \ge \gamma/(2\gamma+1) \approx 0.185$.
It may therefore be concluded that when $\omega \in
\left[\tfrac{\gamma}{(2\gamma+1)}, \frac{1}{2}\right]$  maximising $ \WEI(\bx,
\omega)$ results in the next location for expensive evaluation lying in the
Pareto set of available solutions. However, this is not guaranteed for
other values of $\omega$.   These results are illustrated in 
Figure~\ref{fig:wei_contours}, which shows $\WEI$ as a function of $\mu-\fstar$
and $\sigma$ for $\omega = 0, 0.1$ and $1$; cf Figure~\ref{fig:infill-contour} 
for $\omega = 0.5$.   The complicated nature of
$\WEI$ is apparent when $\omega = 0.1$.  When $\omega
= 0$ the acquisition function might be expected to yield purely
exploratory behaviour.  However, in this case although locations with high
variance are preferred over those with low variance \textit{with the same
  $\mu$}, the acquisition function guides the search towards locations with
high variance but a mean prediction close to $\fstar$.  Purely exploitative
behaviour might be expected when $\omega = 1$.  In this case the
acquisition function is maximised for large $\mu$ and small $\sigma$, which
implies that the location with the smaller $\sigma$ will be preferred from two
locations with the same $\mu$.  Consequently, although the acquisition
function in this case encourages exploitation (preferring large $\mu$) it
discourages exploration (preferring small $\sigma$).  This is in contrast
to standard EI ($\omega = 0.5$, Figure.~\ref{fig:infill-contour})
which  prefers the high variance, more exploratory, location from two locations with the same $\mu$.

\subsubsection{Probability of Improvement.}
\label{sec:prob-impr}

The Probability of Improvement (PI) is one of the earliest proposed infill
criteria \citep{kushner:ego}. It is the probability that the prediction at
a location $\bx$ is greater than the best observed (expensively evaluated)
function value $\fstar$. As the predictive
distribution  is Gaussian, PI may be calculated in closed form:
\begin{equation}
\label{eq:pi}
\PI(\bx) = p(f > \fstar \given \bx, \mD) = \Phi(s(\bx)).
\end{equation}
Thus $\PI(\bx)$ is  the
volume of the predictive distribution lying above  $\fstar$.

Since,
\begin{equation}
  \frac{\partial \PI}{\partial \mu} = \frac{1}{\sigma(\bx)} \phi(s(\bx))
\end{equation}
is positive for all $\mu(\bx)$ and $\sigma(\bx)$, PI is monotonically
increasing with increasing mean prediction for fixed uncertainty. Thus,
 as might be expected, at fixed uncertainty, locations where
the mean is predicted to be large are preferred. Interestingly, as 
Figure~\ref{fig:infill-contour} illustrates, such a straightforward monotonic
relationship does not exist with respect to uncertainty as shown by
\begin{equation}
\frac{\partial \PI}{\partial \sigma} = -s(\bx) \phi(s(\bx)).
\label{eq:dPIdsigma}
\end{equation}

When the improvement in the mean is negative $s(\bx) < 0$
then~\eqref{eq:dPIdsigma} shows that PI increases with uncertainty $\sigma$.
However, in contrast to EI and UCB,  when $\mu(\bx) > \fstar$ 
then~\eqref{eq:dPIdsigma} shows that PI decreases with  $\sigma$; tha
locations with small uncertainty are preferred to those with high uncertainty.  
Therefore, the
location $\xnext$ selected by PI is not guaranteed to be a member of the
maximal non-dominated set of candidates. In other words, there may be
candidate locations $\bx''$ which are more exploratory
($\sigma(\bx'') > \sigma(\xnext)$) while having the same mean prediction
$(\mu(\bx'') = \mu(\xnext))$ as the $\xnext$ selected by PI.

In practice, such behaviour leads to an overly exploitative scheme, see for
example \citep{jones:taxonomy}. To combat this exploitative nature, usually
a higher target than the best observed value, $\fstar$, is set for
computing the probability of improvement. This often improves the performance of
PI-based BO \citep{jones:taxonomy, kushner:ego, lizotte:practicalBO}. 
As Figure~\ref{fig:infill-contour} shows, this can be
attributed to the fact that solutions are evaluated as if their improvement
were negative where the PI criterion encourages exploration as well as
exploitation. Although this modification tends to improve performance,
there is, however, no natural choice for a suitable high target.

\subsection{Exploration and Exploitation Trade-off}
\label{sec:expl-expl-trade}

As discussed above, the EI and UCB infill criteria select the next location
to be expensively evaluated as one of the locations that are members of the
maximal non-dominated set of available locations, namely 
$\mP$~\eqref{eq:pareto-set}, the Pareto set resulting from simultaneous maximisation
of $\mu(\bx)$ and $\sigma(\bx)$. PI only selects from $\mP$ when 
$\mu(\bx) < \fstar$ and in practice an artificially high $\fstar$ is used
to promote exploration. Note, however, that EI and UCB select from different
regions of the Pareto set, balancing exploitation and exploration differently.
Indeed, the  proof of convergence for BO with UCB relies on varying the selection position along the Pareto
front as the optimisation proceeds, becoming more exploratory in later stages
\citep{srinivas:ucb}.

Inspection of Figure~\ref{fig:infill-contour} shows that EI is more 
exploitative than UCB in the sense that if the solutions available for
selection all have the same upper confidence bound, that is they lie on a
contour of $\UCB$, then maximising $\EI$ will choose the most exploitative
of them.  Conversely, if the available solutions all have the same EI, then
maximising $\UCB$ will choose the most exploratory.

\section{Utilising the Exploration vs. Exploitation trade-off front}
Previous works \citep{bischl:mbo, grobler:simple, feng:egomo} have used the
exploration vs. exploitation (EE) front in a batch setting, in which  
multiple locations in the EE Pareto set are selected  to be evaluated in parallel. 
\citet{feng:egomo} use the two weighted components of $\WEI$~\eqref{eq:ei-omega}
as the two objectives defining a trade-off front that is approximated via the use of
a multi-objective evolutionary algorithm (MOEA). They select batches of $q$ 
solutions to be expensively evaluated in parallel by choosing the two
extremal solutions of the approximated Pareto set and the remaining $q-2$ locations equally
spread (in objective space) across the set. 
\citet{grobler:simple} replace the $\WEI$ formulation with a trade-off front
consisting of the surrogate model's mean and variance, again using a 
MOEA to approximate the Pareto set. They select a batch of locations 
consisting of the two extremal solutions of the set, together with the location that maximises 
EI, and equally spaced solutions across the set.
\citet{bischl:mbo} consider the maximisation of an additional objective, namely the decision
space distance to each solution's nearest
neighbour, thus promoting exploration.  They also limit the size of the MOEA
population to be the batch size in order to avoid the problem of explicitly selecting a batch of 
locations from a large Pareto set.

The use of the EE front in the sequential setting is much less explored.
However, \citet{zilinskas:tradeoff} have recently highlighted the importance of
visualising of the EE front to better inform model selection and they recommend that future
researchers should aim to exploit the EE front further.

Here we focus on the sequential BO framework (recall
Algorithm~\ref{alg:bo}) and consider algorithms that select the next
location for expensive evaluation from the entire Pareto set of feasible
locations. Use of an efficient evolutionary multi-objective search
algorithm means that finding an approximation $\Papprox$ to $\mP$ has about
the same computational expense as maximising a scalar acquisition function
such as EI or UCB directly. In this work the approximate Pareto set of
model predictions is found using a standard evolutionary optimiser, NSGA-II
\citep{deb01NSGAII}.

We note that while proofs of convergence for particular trade-offs between
exploration and exploitation exist \citep{bull:convergence, srinivas:ucb},
it is clear that merely selecting locations for any fixed
exploration-exploitation weighting are not guaranteed to converge. At the
two extremes, purely exploitative schemes select
$\xnext = \argmax_{\bx \in \mX} \mu(\bx)$ and purely exploratory schemes
select $\xnext = \argmax_{\bx \in \mX} \sigma(\bx)$. The former are liable
to become stuck at local optima, while the latter visits each location with
the maximum posterior variance $\sigma^2(\bx)$, thus reducing the
uncertainty of the model, as quantified by the entropy of the predictive
posterior. This will lead to the eventual location of the optimum, but only
very slowly as even very unpromising locations where $\mu(\bx) \ll \fstar$
are visited.

In Section~\ref{sec:experiments} we evaluate the performance
of the purely exploitative and exploratory strategies, denoted
\textit{Exploit} and \textit{Explore} respectively. Since all solutions in
the Pareto set may be considered equally good and dominate all other
feasible locations, we also consider the \textit{PFRandom} algorithm, which
selects a solution at random from $\Papprox$ for the next expensive
evaluation.

As discussed above, the maximally exploratory strategy will converge to the
global optimum, but very slowly. At the other extreme of the Pareto front,
a greedy, exploitative, strategy, while converging quickly, risks becoming
stuck at a local optimum.  In the next section, therefore, we seek to
capitalise on the rapid convergence of the exploitative strategy while
avoiding local minima by making occasional exploratory  moves.

\subsection{\texorpdfstring{$\epsilon$-Greedy}{epsilon-Greedy} Bayesian Optimisation}
\label{sec:epsil-greedy-appr}

\begin{algorithm}[t]
  \caption{$\epsilon$-greedy acquisition functions.}
  \label{alg:e-greedy}
    \begin{minipage}[t]{0.48\textwidth}
      \subcaption{\eFront: Pareto front selection.}
      \label{alg:e-greedy-pf-selection}
    \begin{algorithmic}[1]
        \If{ $\mathtt{rand()} < \epsilon  $}
        \State $\Papprox \gets \mathtt{MOOptimise}_{\bx\in\mX}(\mu(\bx), \sigma(\bx))$
        \State{$\xnext \gets \mathtt{randomChoice}(\Papprox)$}
        \Else
        \State{$\xnext \gets \argmax_{\bx \in \mX} \mu(\bx)$}
        \EndIf
      \end{algorithmic}
    \end{minipage}
    \hfill\vline\hfill
    \begin{minipage}[t]{0.48\textwidth}
      \subcaption{\eRandom: Random selection from feasible space.}
      \label{alg:e-greedy-rand-selection}
        \begin{algorithmic}[1]
          \If{ $\mathtt{rand()} < \epsilon  $}
          \State{$\xnext \gets \mathtt{randomChoice}(\mathcal{X})$}
          \Else
          \State{$\xnext \gets \argmax_{\bx \in \mX} \mu(\bx)$}
          \EndIf
        \end{algorithmic}
       \end{minipage}
    \end{algorithm}

Motivated by the success of $\epsilon$-greedy schemes in reinforcement learning
\citep{sutton:rl, tokic:adaptive, minh:rlnature, vanhasselt:deeprl},
we propose two novel BO acquisition functions which use the $\epsilon$-greedy 
methodology to select the next point for expensive evaluation. Both methods
mostly select the most exploitative solution, but differ in which
exploratory solution is selected in a small proportion of steps.

The first method
which we denote \eFront and is summarised in 
Algorithm~\ref{alg:e-greedy-pf-selection}, usually selects the location 
$\xnext$ with the most promising mean prediction from the surrogate model. In
the remaining cases, with probability $\epsilon$, it selects a random location
from the approximate Pareto set $\Papprox$, thus usually selecting a more
exploratory $\xnext$ instead of the most exploitative location available.
The function $\mathtt{MOOptimise}$ denotes the use of a multi-objective
optimiser to generate $\Papprox$. This acquisition function replaces 
line~\ref{choose-xnext} in standard BO, Algorithm~\ref{alg:bo}.

The \eRandom scheme, summarised in 
Algorithm~\ref{alg:e-greedy-rand-selection}, also usually selects 
$\xnext$ with the most promising mean prediction from the surrogate. However,
with probability $\epsilon$ a location is randomly selected (hence
the abbreviation \eRandom) from the entire
feasible space $\mathcal{X}$. 
Selection of $\xnext$ from $\Papprox$
(\eFront, Algorithm~\ref{alg:e-greedy-pf-selection}) might be expected to be more
effective than selecting $\xnext$ from the entire feasible space
(\eRandom, Algorithm~\ref{alg:e-greedy-rand-selection}) because a selection from $\mX$
is likely to be dominated by $\Papprox$ and therefore is likely to be less
exploratory and less exploitative. 

We remark that these $\epsilon$-greedy schemes are different to that proposed
by \citet{bull:convergence}, which greedily selects the location with
maximum expected improvement with probability $1-\epsilon$, and randomly
chooses a location the remainder of the time. This is different from our
proposals because the Bull scheme greedily maximises EI rather than
exploitation ($\mu$).

\section{Experimental Evaluation}
\label{sec:experiments}

\begin{table}[t]
    \centering
    \setlength{\tabcolsep}{4pt}
  \begin{tabular}[t]{llr}
  \toprule
  \bfseries Name & \bfseries Domain &  $d$ \\
  \midrule
  WangFreitas \citep{wangfreitas:2014}
          & $[0, 1]$  & 1\\
  Branin$^\dagger$
          & $[-5, 0]\times[10, 15]$ & 2 \\
  BraninForrester \citep{forrester:2008}
          & $[-5, 0]\times[10, 15]$ & 2 \\
  Cosines \citep{gonzalez:2016a}
          & $[0, 0]\times[5, 5]$ & 2\\
  logGoldsteinPrice$^\dagger$
          & $[-2, -2]\times[2, 2]$ & 2\\
  logSixHumpCamel$^\dagger$
          & $[-3, 2]\times[3, 2]$ & 2\\
  logHartmann6$^\dagger$
          & $[0, 1]^d$ & 6\\
  logGSobol \citep{gonzalez:2016b}
          & $[-5, 5]^d$ & 10\\
  logRosenbrock$^\dagger$
          & $[-5, 10]^d$ & 10\\
  logStyblinskiTang$^\dagger$
          & $[-5, 5]^d$ & 10\\
  \bottomrule
  \end{tabular}
  \caption{Functions used in these experiments, along with their domain and 
  dimensionality, $d$. Formulae can be found as cited or at 
  \url{http://www.sfu.ca/~ssurjano/optimization.html} for those labelled with $\dagger$.
  Full details of all evaluated functions can also be found in the supplementary
  material.}
  \label{tbl:function_details}
  \end{table}

We investigate the performance of the two proposed $\epsilon$-greedy methods,
\eFront and \eRandom, by evaluating them on ten benchmark functions with a 
range of domain sizes and dimensionality; see Table~\ref{tbl:function_details}
for details. Note that the benchmarks are couched as minimisation problems. 
In common with other works \citep{schonlau:ego,jones:ego, wang:niching,
wagner:transformations}, the functions prefixed with \textit{log} are 
log-transformed, \ie the logarithm of each observed values $\log(f(\bx))$ is
modelled rather than observed value $f(\bx)$ itself. Where the observations can
be negative, a  constant larger than minimum value of the function
is added.\footnote{We have used prior information on the function's minimum
value to choose the constant, but the actual value is immaterial because
the function observations are in any case standardised as part of the GP modelling.} 
These functions are transformed in this \textit{grey-box} fashion, using a
small amount of prior information about the scales of the function, because we want
the surrogate model to be as accurate as possible. As discussed in the seminal
work of \citet{jones:ego}, it is often possible to improve poorer surrogate
model fits, as one typically observes with the untransformed functions, by
using the log transformation. The equations defining each transformed function and
optimisation results of all methods on the \textit{untransformed} functions are
available in the supplementary material.  We discuss  the differences in
optimisation performance between the standard and log-transformed
functions below. 

We compare the two proposed methods to the purely exploitative
and exploratory strategies, denoted \textit{Exploit} and \textit{Explore} 
respectively, as well as random selection from the approximated Pareto
front, \textit{PFRandom}. Their performance is also compared to the infill 
criteria discussed in Section~\ref{sec:infill-criteria}, namely Expected 
Improvement (EI), Upper Confidence Bound (UCB) and Probability of 
Improvement (PI). In addition, we compare the performance of all the infill 
criteria with the quasi-random search produced by max-min Latin Hypercube 
Sampling (LHS, \citep{mckay:lhs}). LHS is the generalisation of a Latin square,
in which samples are placed in rows and columns of a square such that each
sample resides in its own row and column. The max-min variant of LHS tries to
maximise the minimum distance between each sample.

The methods were evaluated on the synthetic benchmark functions in 
Table~\ref{tbl:function_details}, with a budget of 250 function evaluations
that included $M=2d$ initial LHS samples (Algorithm~\ref{alg:bo},
Line~\ref{alg-bo-lhs}). To allow statistical performance measures to be
used, each optimisation was repeated 51 times.   The same 
sets of initial samples were used for each method's runs to allow for paired 
statistical comparisons between the methods to be carried out.  In
all experiments a value of $\epsilon = 0.1$ was used for both \eFront and 
\eRandom. The UCB algorithm was run with $\beta_t$ adjusted according to
the schedule  defined for 
continuous functions in Theorem 2 of \citet{srinivas:ucb} with $a = b = 1$
and $\delta = 0.01$.
All acquisition functions were optimised with NSGA-II \citep{deb01NSGAII},
apart from PI which was optimised following the common practise
\citep{gpyopt, botorch} of uniformly sampling $\mX$ and optimising the 10 most
promising locations with L-BFGS-B \citep{byrd:lbfgs}. 
In both cases the optimisation budget was $5000d$
evaluations. The multi-start strategy was used to optimise PI because,
as shown in Section~\ref{sec:prob-impr}, the maximiser of PI may not lie in the
Pareto set of $\mu(\bx)$ and $\sigma(\bx)$. For NSGA-II, we set the 
parameters to commonly used values: the population size was $100d$, the number
of generations was $50$ (100 generations lead to no significant improvement in
performance), the crossover and mutation probabilities were $0.8$ 
and $\tfrac{1}{d}$ respectively, and both the distribution indices for 
crossover and mutation were $20$.

\begin{table}[t]
\setlength{\tabcolsep}{2pt}
\sisetup{table-format=1.2e-1,table-number-alignment=center}
\resizebox{\textwidth}{!}{%
\begin{tabular}{l SS SS SS SS SS}
\toprule
\bfseries Method
    & \multicolumn{2}{c}{\bfseries WangFreitas (1)} 
    & \multicolumn{2}{c}{\bfseries BraninForrester (2)} 
    & \multicolumn{2}{c}{\bfseries Branin (2)} 
    & \multicolumn{2}{c}{\bfseries Cosines (2)} 
    & \multicolumn{2}{c}{\bfseries logGoldsteinPrice (2)} \\ 
    & \multicolumn{1}{c}{Median} & \multicolumn{1}{c}{MAD}
    & \multicolumn{1}{c}{Median} & \multicolumn{1}{c}{MAD}
    & \multicolumn{1}{c}{Median} & \multicolumn{1}{c}{MAD}
    & \multicolumn{1}{c}{Median} & \multicolumn{1}{c}{MAD}
    & \multicolumn{1}{c}{Median} & \multicolumn{1}{c}{MAD}  \\ \midrule
    LHS & 1.27e-02 & 1.80e-02 & 4.59e-01 & 4.73e-01 & 1.31e-01 & 1.33e-01 & 4.79e-01 & 2.71e-01 & 1.08e+00 & 7.69e-01 \\
    Explore & 1.04e-02 & 1.42e-02 & 4.58e-01 & 3.52e-01 & 1.66e-01 & 1.56e-01 & 4.56e-01 & 2.20e-01 & 1.01e+00 & 5.50e-01 \\
    EI & 2.00e+00 & 6.91e-11 & \statsimilar 2.47e-06 & \statsimilar 3.23e-06 & \statsimilar 4.15e-06 & \statsimilar 3.76e-06 & \statsimilar 6.31e-06 & \statsimilar 7.68e-06 & 2.73e-06 & 3.34e-06 \\
    PI & 2.06e+00 & 8.24e-02 & 3.73e-04 & 3.70e-04 & 2.26e-05 & 3.22e-05 & \statsimilar 2.50e-03 & \statsimilar 3.18e-03 & 2.92e-03 & 4.32e-03 \\
    UCB & 2.00e+00 & 1.26e-11 & \statsimilar 4.96e-06 & \statsimilar 6.22e-06 & \statsimilar 4.42e-06 & \statsimilar 4.06e-06 & \statsimilar 7.12e-06 & \statsimilar 8.86e-06 & 6.15e-06 & 6.17e-06 \\
    PFRandom & \statsimilar 2.00e-04 & \statsimilar 2.96e-04 & 2.70e-03 & 3.65e-03 & 1.67e-03 & 2.17e-03 & \statsimilar 8.82e-03 & \statsimilar 1.14e-02 & 2.54e-03 & 3.31e-03 \\
    \eRandom & \best 1.04e-06 & \best 1.54e-06 & \best 2.00e-06 & \best 2.49e-06 & \statsimilar 3.17e-06 & \statsimilar 2.46e-06 & 8.66e-06 & 1.21e-05 & 2.33e-06 & 2.36e-06 \\
    \eFront & \statsimilar 2.00e+00 & \statsimilar 3.72e-11 & \statsimilar 2.31e-06 & \statsimilar 3.01e-06 & \statsimilar 3.57e-06 & \statsimilar 3.13e-06 & \best 2.02e-06 & \best 2.52e-06 & \best 8.76e-07 & \best 1.08e-06 \\
    Exploit & 2.00e+00 & 6.00e-09 & 4.61e-06 & 6.04e-06 & \best 3.08e-06 & \best 3.29e-06 & 4.13e-01 & 6.12e-01 & 2.26e-06 & 2.90e-06 \\
\bottomrule
\toprule
\bfseries Method
    & \multicolumn{2}{c}{\bfseries logSixHumpCamel (2)} 
    & \multicolumn{2}{c}{\bfseries logHartmann6 (6)} 
    & \multicolumn{2}{c}{\bfseries logGSobol (10)} 
    & \multicolumn{2}{c}{\bfseries logRosenbrock (10)} 
    & \multicolumn{2}{c}{\bfseries logStyblinskiTang (10)} \\ 
    & \multicolumn{1}{c}{Median} & \multicolumn{1}{c}{MAD}
    & \multicolumn{1}{c}{Median} & \multicolumn{1}{c}{MAD}
    & \multicolumn{1}{c}{Median} & \multicolumn{1}{c}{MAD}
    & \multicolumn{1}{c}{Median} & \multicolumn{1}{c}{MAD}
    & \multicolumn{1}{c}{Median} & \multicolumn{1}{c}{MAD}  \\ \midrule
    LHS & 6.52e+00 & 1.10e+00 & 3.37e-01 & 1.10e-01 & 1.51e+01 & 9.03e-01 & 1.16e+01 & 5.39e-01 & 2.85e+00 & 1.77e-01 \\
    Explore & 6.53e+00 & 1.24e+00 & 3.07e-01 & 6.85e-02 & 1.75e+01 & 1.42e+00 & 1.28e+01 & 4.82e-01 & 3.19e+00 & 1.13e-01 \\
    EI & 7.42e-05 & 9.19e-05 & \statsimilar 1.06e-03 & \statsimilar 6.73e-04 & 7.15e+00 & 1.58e+00 & 6.62e+00 & 6.58e-01 & 2.34e+00 & 2.79e-01 \\
    PI & 1.46e-01 & 1.58e-01 & \statsimilar 6.15e-04 & \statsimilar 7.69e-04 & 6.29e+00 & 1.61e+00 & 6.89e+00 & 9.49e-01 & 2.29e+00 & 2.37e-01 \\
    UCB & 3.84e+00 & 1.36e+00 & 2.04e-01 & 3.21e-02 & 1.45e+01 & 6.16e-01 & 8.31e+00 & 5.90e-01 & 3.19e+00 & 1.13e-01 \\
    PFRandom & 1.52e-01 & 1.52e-01 & 6.57e-02 & 3.27e-02 & 5.60e+00 & 1.73e+00 & 5.23e+00 & 4.98e-01 & 2.70e+00 & 3.15e-01 \\
    \eRandom & \best 3.81e-05 & \best 2.96e-05 & \best 5.09e-04 & \best 3.59e-04 & \statsimilar 5.13e+00 & \statsimilar 1.86e+00 & \statsimilar 4.75e+00 & \statsimilar 7.85e-01 & \statsimilar 1.61e+00 & \statsimilar 3.12e-01 \\
    \eFront & \statsimilar 4.06e-05 & \statsimilar 4.66e-05 & 7.71e-04 & 4.82e-04 & \best 5.06e+00 & \best 1.37e+00 & \statsimilar 4.64e+00 & \statsimilar 6.25e-01 & \best 1.53e+00 & \best 4.49e-01 \\
    Exploit & \statsimilar 4.21e-05 & \statsimilar 4.95e-05 & \statsimilar 6.37e-04 & \statsimilar 5.82e-04 & \statsimilar 5.27e+00 & \statsimilar 1.60e+00 & \best 4.54e+00 & \best 6.19e-01 & 1.82e+00 & 3.71e-01 \\
\bottomrule
\end{tabular}
}
\caption{Median absolute distance (\textit{left}) and median absolute deviation from the
median (MAD, \textit{right}) from the optimum after 250 function evaluations across 
the 51 runs. The method with the lowest median performance is shown
in dark grey, with those with statistically equivalent performance
are shown in light grey. }
\label{tbl:synthetic_results}
\end{table}

Table~\ref{tbl:synthetic_results} shows the median regret, \ie the median
difference between the estimated optimum $\fstar$ and the true optimum %
over the 51 repeated experiments, together with the median
absolute deviation from the median (MAD). The method with the lowest (best)
median regret on each function is highlighted in dark grey, and those which are
statistically equivalent to the best method according to a one-sided paired 
Wilcoxon signed-rank test \citep{knowles:testing} with Holm-Bonferroni correction
\citep{holm:test} ($p\geq0.05$), are shown in light grey.

\begin{figure}[t]
\includegraphics[width=\textwidth]{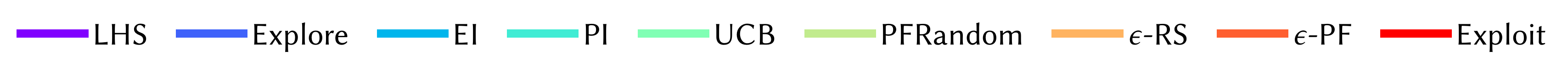}\\
\includegraphics[width=\textwidth, clip, trim={7 0 8 0}]%
{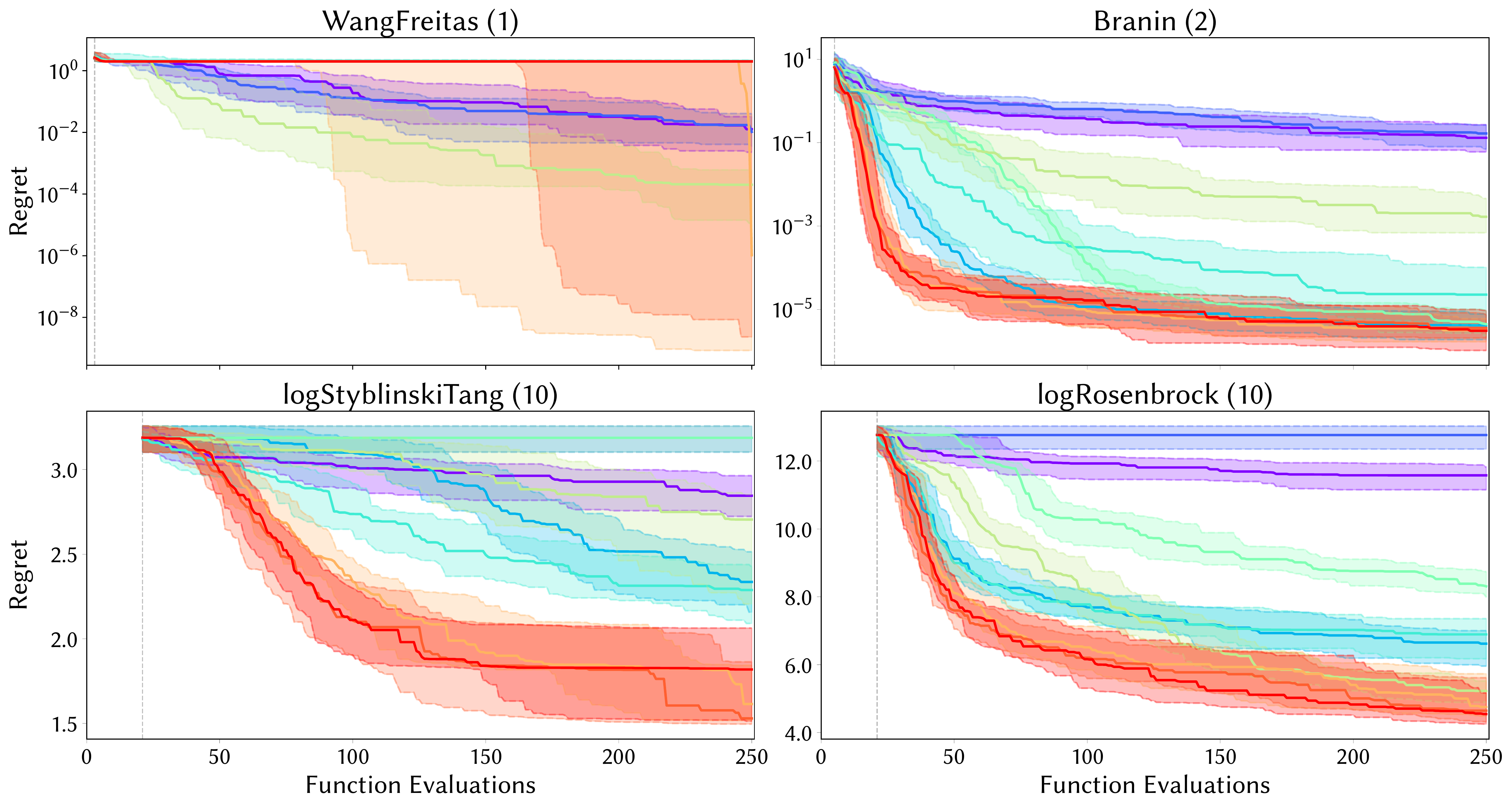}
\caption{Illustrative convergence plots for four benchmark problems.  Each
  plot shows the median difference between the best function value seen and
  the true optimum (regret), with shading representing the interquartile range across 
  the 51 runs. The dashed vertical line indicates the end of the initial LHS 
  phase.}
\label{fig:conv_plots}
\end{figure}

Figure~\ref{fig:conv_plots} shows the 
convergence of the various algorithms on  four illustrative test problems in 
$d = 1, 2$ and $10$ dimensions. Convergence plots for all the benchmark 
problems are available in the supplementary material, and Python code to
generate figures and reproduce all experiments is available
online\footnote{\url{https://github.com/georgedeath/egreedy/}}.

As might be expected, Latin Hypercube Sampling (LHS) and purely exploratory
search (Explore), which have roughly equivalent performance, are not the best 
methods on any of the test problems.

Perhaps surprisingly, none of the three well-known acquisition functions, 
EI, UCB and PI, has the best median performance after 250 evaluations, although
all three are statistically equivalent to the best method on $d=2$ Cosines, 
and EI and UCB have good performance on the $d=2$ Branin and 
BraninForrester problems. 
In contrast, the $\epsilon$-greedy algorithms
\eFront and \eRandom perform well across the range of problems,
particularly on the higher-dimensional problems. Interestingly, Exploit,
which always samples from the best mean surrogate prediction is competitive
for most of the high dimensional problems. This indicates one of the main
conclusions of this work, namely that as the dimension of decision space
increases the approximate modelling of $f(\bx)$ is so poor that even
adopting the modelled most-exploitative solution inherently leads to some
(unintended) exploration.

While pure exploitation combined with fortuitous exploration appears to be
a good strategy for many problems, introducing some deliberate exploration
can be important. This is particularly apparent on the WangFreitas problem
\citep{wangfreitas:2014} which contains a large local optimum and a narrow
global optimum that is surrounded by plateaux; see supplementary material
for a plot. Convergence on this problem is shown in
Figure~\ref{fig:conv_plots}, which demonstrates how LHS sampling and a
purely exploratory strategy (Explore) converge slowly towards the optimum,
while Exploit fails to find the vicinity of the optimum in any case. On the
other hand, the deliberate exploratory moves incorporated in both
$\epsilon$-greedy methods and PFRandom (random selection from the Pareto
set) enable some of the runs to converge to the optimum. The \eRandom method, 
which makes exploratory moves from the entire feasible space, is most 
effective, although as discussed below, generally we find \eFront to be more 
effective.

\begin{figure}[t]
\centering
\includegraphics[width=\textwidth, clip, trim={7 8 7 8}]
{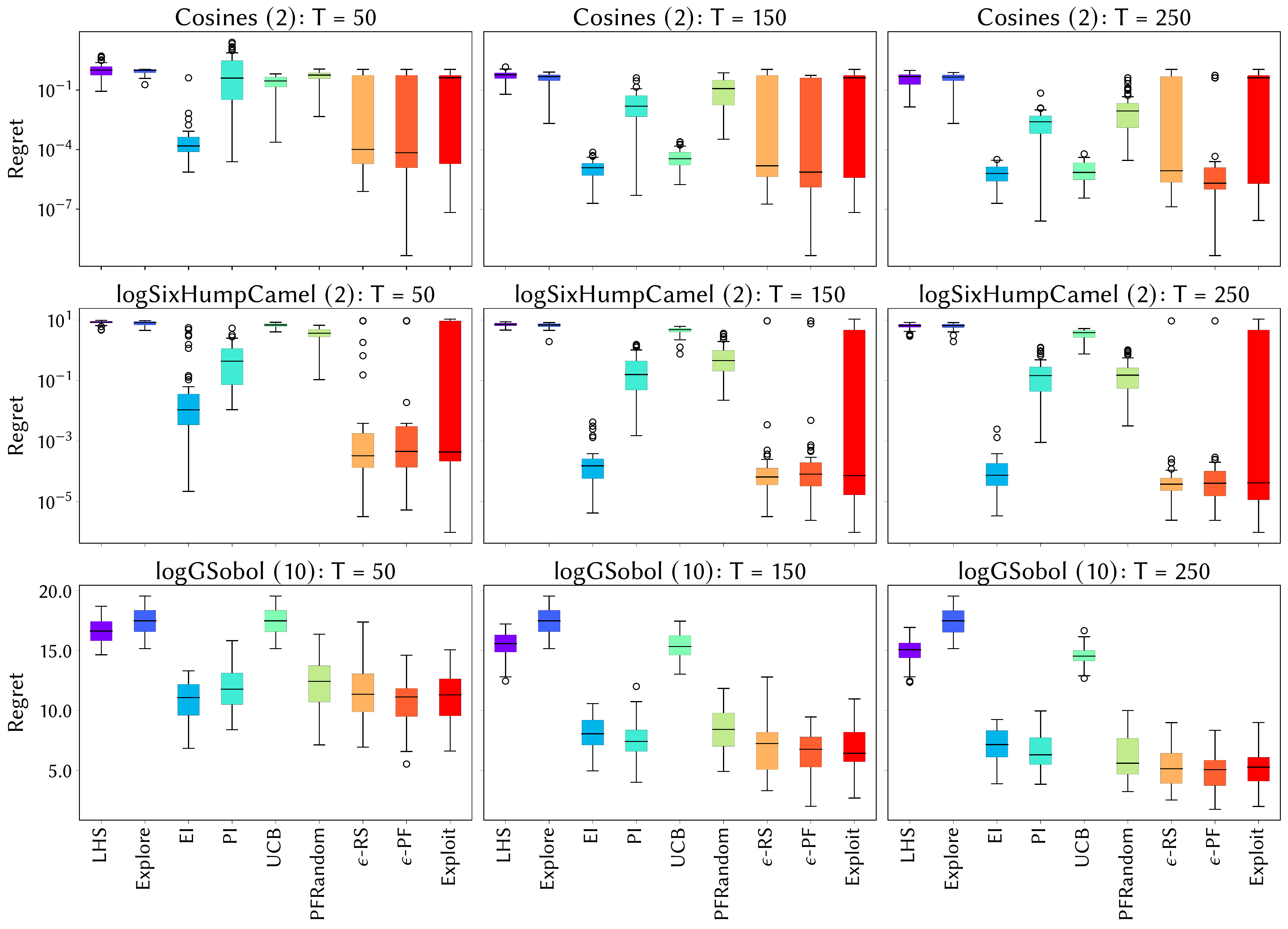}
\caption{Distribution of the best-seen function values after 50 (\textit{left}), 
150 (\textit{centre}) and 250 (\textit{right}) function evaluations on three benchmark problems.}
\label{fig:box_plots}
\end{figure}
Figure~\ref{fig:box_plots} shows the distribution of the best-seen function 
evaluations for each of the evaluated algorithms on three benchmark problems 
for budgets of $T = 50$, $150$ and $250$ function evaluations. Again, we see
in the two-dimensional Cosines and logSixHumpCamel plots that driving 
the optimisation process solely by exploiting the surrogate's mean
prediction can fail to correctly identify the optimum because the model is
inaccurate and may miss, for example, a small scale optimum. When 
$f$ is modelled poorly, then the mean function will not accurately represent 
the true function. However, as is the case with the logGSobol plot and indeed
the other ten-dimensional functions, pure exploitation
can provide a sufficient driver for optimisation, because the inaccurate
and changing surrogate (as new evaluations become available) induces
sufficient exploration.  We note however, that the $\epsilon$-greedy
algorithms, incorporating deliberate exploration, offer more consistent
performance. 

A common trend apparent across the both Figures~\ref{fig:conv_plots} 
and~\ref{fig:box_plots} is that EI tends to initially improve at a slower rate than
the two $\epsilon$-greedy methods, but then catches up to a greater or lesser
extent after more function evaluations. This is well illustrated in the
logSixHumpCamel plot in Figure~\ref{fig:box_plots} and also in the Branin and 
logRosenbrock plots in Figure~\ref{fig:conv_plots}. UCB performs poorly on the
higher dimensional functions. This may be due to the value of $\beta_t$ 
used, as the convergence proofs in \citep{srinivas:ucb} rely on $\beta_t$ 
increasing with the dimensionality of the problem, leading to over-exploration.
One may argue that this can be overcome by simply using a smaller $\beta_t$ 
value, set in some \textit{ad hoc} manner. However, with no \textit{a priori}
knowledge as to how to select the parameter on a per-problem basis, we suggest
that this is not a feasible strategy in practice.

\subsubsection*{How greedy? Choosing  $\epsilon$}
\begin{figure}[t]
\centering
\includegraphics[width=\textwidth, clip, trim={7 8 0 8}]
{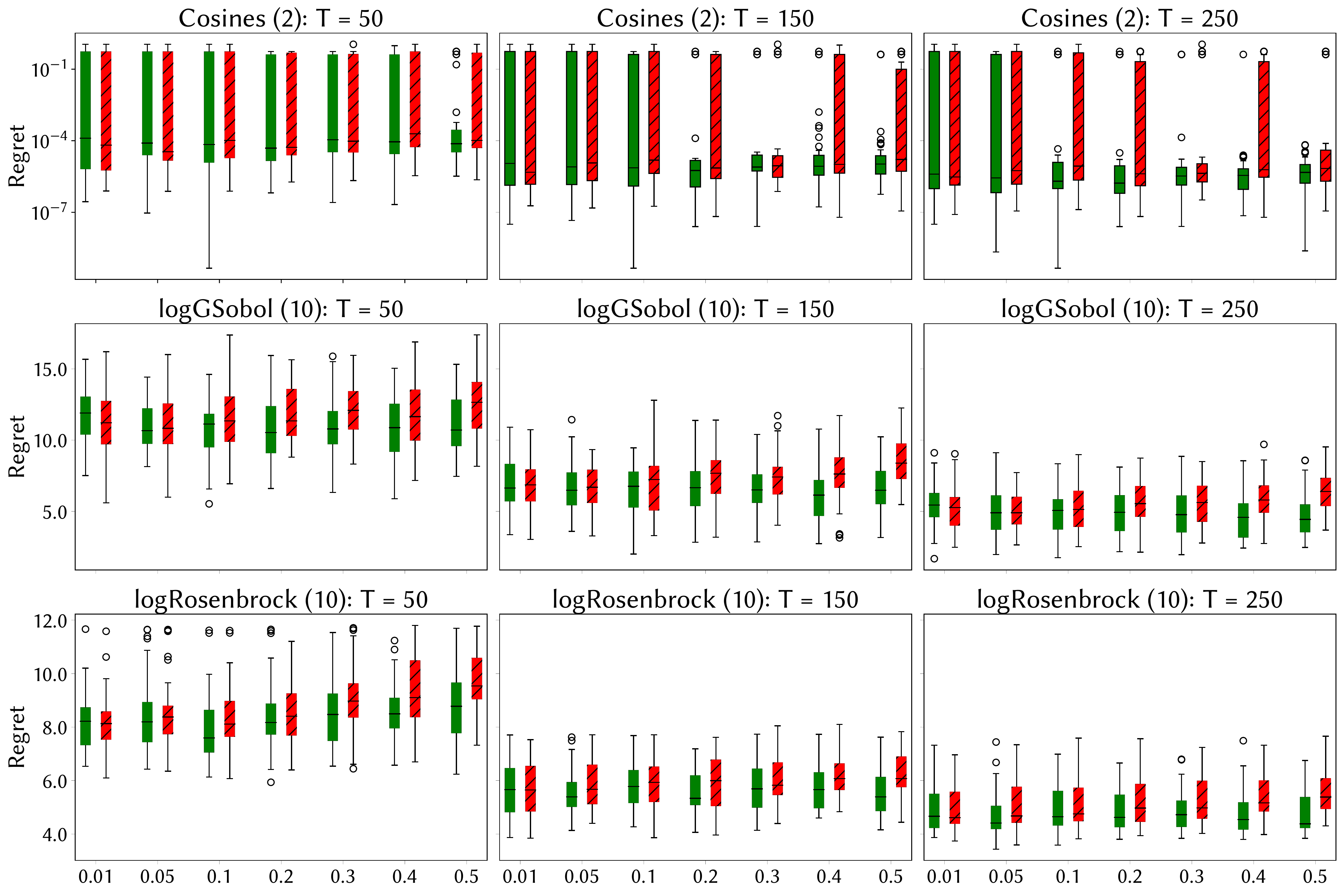}
\caption{Comparison of \textcolor{OliveGreen}{\eFront} 
(\textcolor{OliveGreen}{green}) and \textcolor{Maroon}{\eRandom} 
(\textcolor{Maroon}{red, hatched}) for different values of 
$\epsilon$ (horizontal axis) after 50~(\textit{left}), 150 (\textit{centre})
and 250 (\textit{right}) function evaluations.}
\label{fig:side_by_side_comparison}
\end{figure}
Although the $\epsilon$-greedy algorithms perform well in comparison with
conventional acquisition functions, it is unclear what value of $\epsilon$
to choose, and indeed whether the exploratory moves should choose from the
approximate Pareto front (\eFront) or from the entire feasible space
(\eRandom) which is marginally cheaper.
Figure~\ref{fig:side_by_side_comparison} illustrates the effect of
$\epsilon$ on the performance of \eFront (green) and \eRandom (red,
hatched). As is clear from the Cosines problem, a larger value of
$\epsilon$ may be required to avoid getting stuck because the surrogate is
not modelling the function well enough and needs a larger number of
exploratory samples. 
\begin{figure}[t]
\centering
\includegraphics[width=\textwidth, clip, trim={7 8 0 8}]
{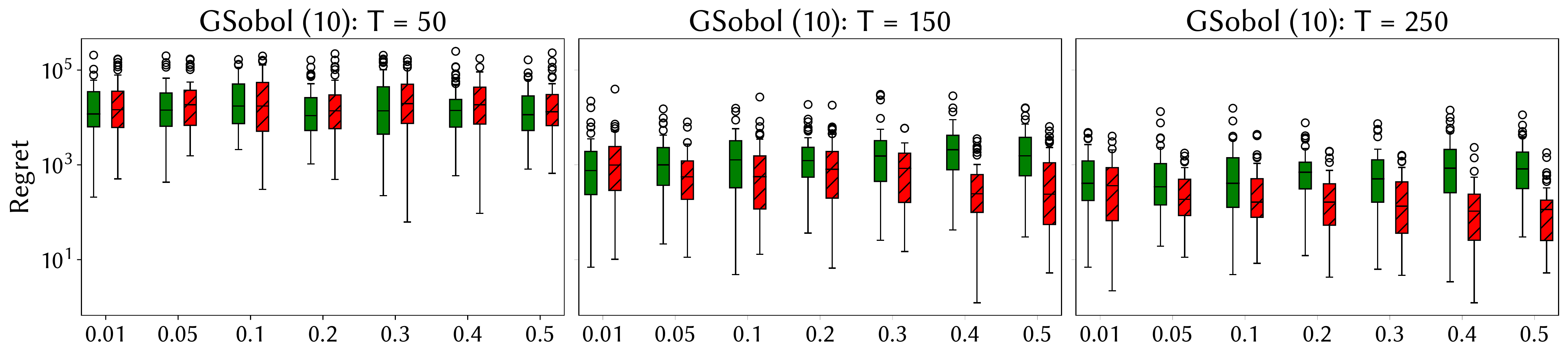}
\caption{A comparison of optimising the GSobol function with
\textcolor{OliveGreen}{\eFront} (\textcolor{OliveGreen}{green}) and 
\textcolor{Maroon}{\eRandom} (\textcolor{Maroon}{red, hatched}) for different
values of $\epsilon$ (horizontal axis) after 50~(\textit{left}), 
150 (\textit{centre}) and 250 (\textit{right}) function evaluations.}
\label{fig:gsobol_unmodified}
\end{figure}
However, there is very little change in performance with $\epsilon$ for the 
higher dimensional decision spaces (\eg logGSobol and logRosenbrock). As
suggested above we attribute this to the inaccurate surrogate modelling in
higher dimensions which leads to a large degree of random search
irrespective of $\epsilon$.

Interestingly, this is not the case for these functions without the log
transformation (Rosenbrock and GSobol). Figure~\ref{fig:gsobol_unmodified} shows the performance
of \eRandom and \eFront for different values of $\epsilon$ on the GSobol 
problem. As can be seen in the figure, increasing $\epsilon$ decreases the
performance of \eFront and increases the performance of \eRandom, in stark
contrast to logGSobol in Figure~\ref{fig:side_by_side_comparison}. This
indicates that the surrogate model is misleading the optimisation because
increasing the frequency of expensively evaluating random locations
and decreasing the frequency of sampling from the Pareto front both
improve optimisation. In this case, the log transformation enables more
accurate modelling of the objective and thus more rapid optimisation. 

Overall, setting $\epsilon=0.1$ appears to be large enough to give good
performance across all problems (see supplementary material for results on
other problems), particularly for the \eFront algorithm.  Larger values
give no real improvement in performance. Empirically it appears that \eFront
gives marginally better performance than \eRandom, as might be expected if
the surrogate describes $f$ well, as is the case in the later stages of
optimisation. In this case, selection from the approximate Pareto front
yields solutions that lie on the maximal trade-off between exploration and
exploitation and may therefore be expected to yield the most information.
However, in cases where the surrogate modelling is particularly poor throughout
the entire optimisation run, as is the case in several of the test problems
without log transformation,
the increased stochasticity provided by \eRandom with larger
values of $\epsilon$ appears useful in overcoming the misleading surrogate
model.

\subsubsection*{Results on the black-box test problems}
Here we briefly describe the optimisation results of the evaluated methods on
the six  test problems without log transformation -- full results are available in the
supplementary material. The \eRandom method is the best performing or
statistically equivalent to the best performing method on all six of the
benchmark problems, with \eFront best or equivalent on five of the six. EI, PI and Exploit
were all the best or equivalent to the best performing on three of the six test
problem. As noted above, \eRandom performs better than
\eFront on the higher dimensional problems, with the two methods giving
equivalent performance on the lower dimensional problems.
The main difference 
of the standard acquisition functions is that performance is closer to that of
the $\epsilon$-greedy methods than on the log-transformed functions. We
attribute this to be a result of poorer surrogate modelling in the presence
of a wide range of objective values so that the $\epsilon$-greedy schemes are less able to exploit the model's mean
predicted value $\mu(\bx)$. We reiterate here, however, that the performance
of both \eRandom and \eFront across the untransformed benchmark functions is
still superior to the standard acquisition functions.

\subsection{Real-World Application: Pipe Shape Optimisation}
\label{sec:real-world-appl}

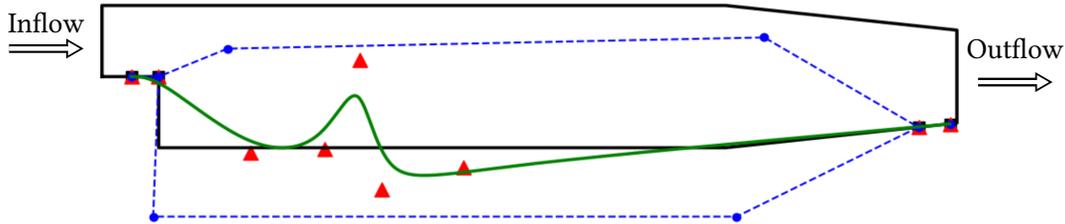
\begin{figure}[t]
  \centering
  \tikzstyle{vecArrow} = [thick, decoration={markings,mark=at position
   1 with {\arrow[semithick]{open triangle 60}}},
   double distance=1.4pt, shorten >= 5.5pt,
   preaction = {decorate},
   postaction = {draw,line width=1.4pt, white,shorten >= 4.5pt}]
  \begin{tikzpicture}[scale=0.9]
    \node[] at (0, 0)
    {\includegraphics[width=0.75\textwidth]{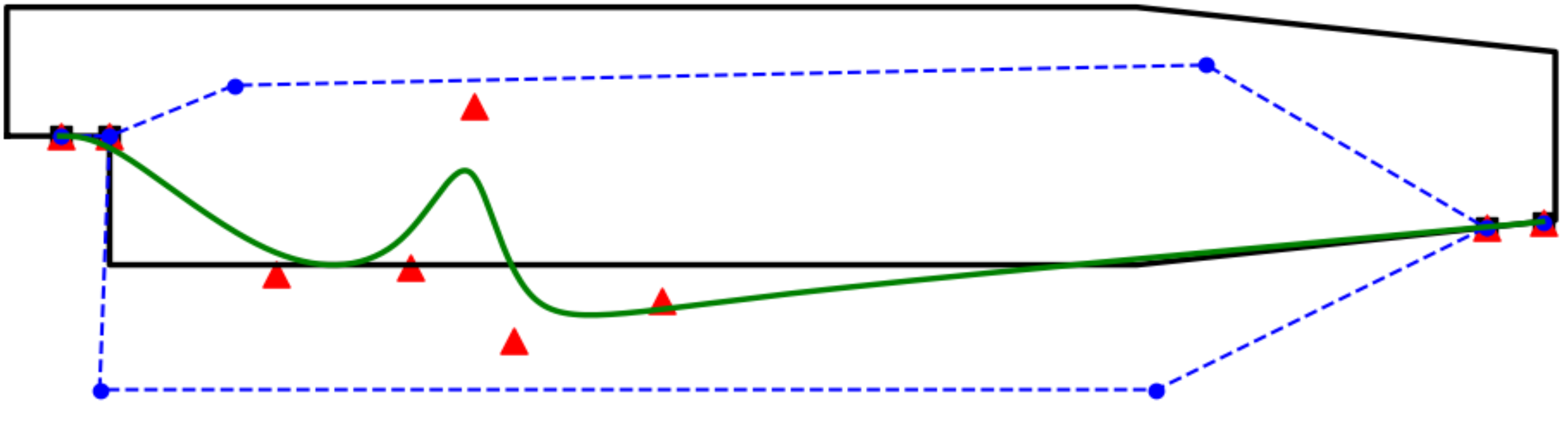}};
    \node[] at (-6.5, 1.3) {Inflow};
    \draw[vecArrow] (-7, 0.95) -- (-6, 0.95);

    \draw[vecArrow] (6, 0.5) -- (7, 0.5);
    \node[] at (6.5, 0.95) {Outflow};
  \end{tikzpicture}
   \centering
  
  \caption{PitzDaily test problem. Fluid enters on the left (Inflow), flows
  through the expanded pipe and leaves on the right (Outflow). The shape of the
  lower boundary is defined by a Catmull-Clark subdivision curve 
  (\textcolor{OliveGreen}{green}) controlled by the locations of control points
  (\textcolor{red}{$\blacktriangle$}). The curve is constrained to lie within 
  the \textcolor{blue}{blue polygon} by penalising the acquisition function for
  solutions that violate it.}
  \label{fig:PitzDaily-geometry}
\end{figure}

We also evaluate the range of acquisition functions on a real-world
computational fluid dynamics optimisation problem. As illustrated in
Figure~\ref{fig:PitzDaily-geometry}, the PitzDaily test problem
\citep{daniels:benchmark} involves optimising the shape of a pipe in order
to reduce the pressure loss between the inflow and outflow. Pressure loss
is caused by a rapid expansion in the pipe (a backward-facing step), which
forces the flow to separate at the edge of the step, creating a
recirculation zone, before the flow re-attaches at some distance beyond the
step. The goal of the optimisation is to discover the shape of the lower
wall of the pipe that minimises the pressure loss, which is evaluated by
running a computational fluid dynamics (CFD) simulation of the
two-dimensional flow. Solution of the partial differential equations
describing the flow means that each function evaluation takes about 60 
seconds --- which is sufficient for us to conduct multiple runs to enable 
statistical comparisons for this problem.

As shown in Figure~\ref{fig:PitzDaily-geometry} and as described in detail
by \citet{daniels:benchmark}, we represent the wall geometry in terms of a
Catmull-Clark sub-division curve, whose control points comprise the
decision variables. Here there are 5 control points, resulting in a
10-dimensional decision vector. The control points are constrained to reside
within a polygon and, therefore, the initial locations used in each optimisation
run are sampled from a uniform distribution, and those that reside outside the
constrained region are discarded and new samples generated to replace them. 
Similarly, the optimisation runs are compared to uniformly sampling 250 
locations rather than Latin hypercube sampling, and are denoted as 
\textit{Uniform} in the following results.

\begin{figure}[t]
\centering
\includegraphics[width=\textwidth]{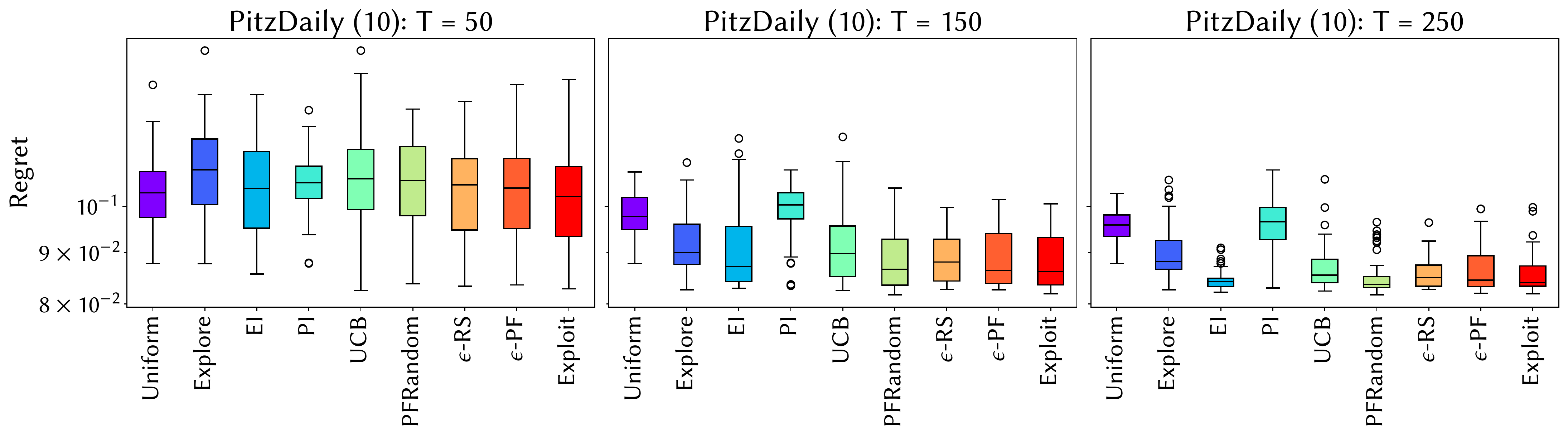}
\caption{Distribution of the best-seen function values after 50 (\textit{left}), 
150 (\textit{centre}) and 250 (\textit{right}) function evaluations on the real-world PitzDaily test problem.}
\label{fig:conv_plot_box_plot_pitzdaily}
\end{figure}

Figure~\ref{fig:conv_plot_box_plot_pitzdaily}, shows random selection from the
Pareto front (PFRandom) had the
best median fitness  after 250 function evaluations,
but EI, \eFront, \eRandom and Exploit were all statistically equivalent. We
remark that the optimum discovered outperforms that discovered by
\citet{nilsson:pitzdaily}. 

\begin{figure}[t]
  \centering
\includegraphics[width=0.86\textwidth]{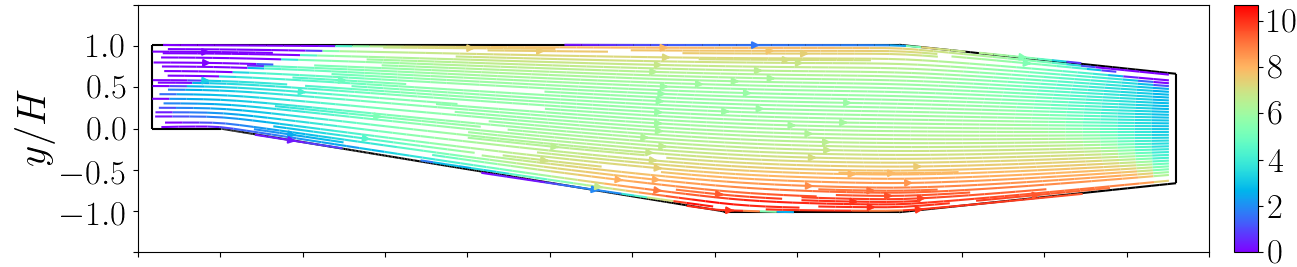}
\includegraphics[width=0.86\textwidth]{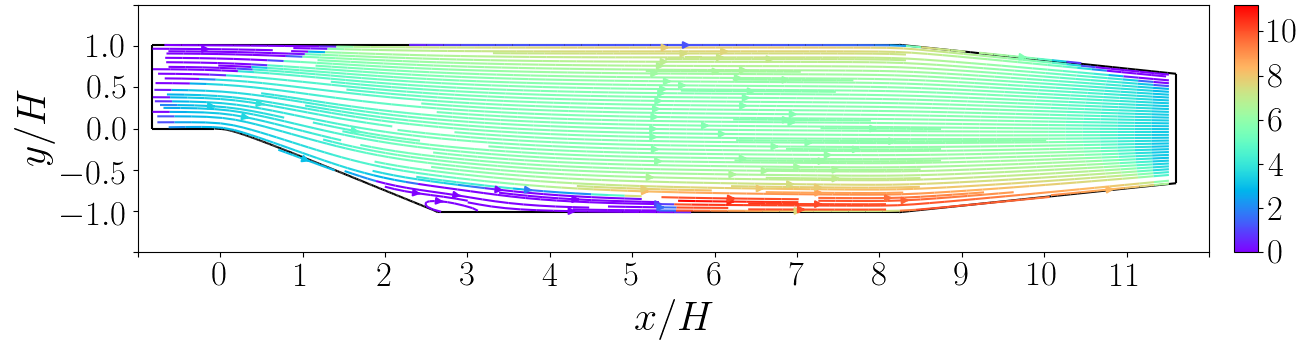}
\caption{The streamlines for two solutions: the local optimum identified by
  \citet{nilsson:pitzdaily} (\textit{upper}) and the best
  estimation of the global optimum from one of the runs using the Bayesian
  optimiser (\textit{lower}). Colour indicates fluid speed (normalised units). Good
  solutions typically replace the backward step with a slope.}
\label{fig:streamlines_pitzdaily}
\end{figure}

We observe that good solutions typically replace the step shown in 
Figure~\ref{fig:PitzDaily-geometry} with a slope, as illustrated by the two
solutions shown in Figure~\ref{fig:streamlines_pitzdaily}. This improves the
performance because it reduces the size of the recirculation zone
immediately following the increase in the tube's width. Generally, the size
of the recirculation zone is reduced for shallower slopes, resulting in a
reduced flow velocity (as the streamlines suggest) and increased frictional
pressure recovery.  However, such a shallow slope that the recirculation
zone is completely removed (as found by an adjoint optimisation method)
does not perform best \citep{nilsson:pitzdaily}. The Bayesian optimiser
consistently discovers a wall shape that results in a small recirculation
zone that more effectively dampens the flow, resulting in a smaller
pressure loss \citep{daniels:diffuser}.

\subsection{Real-World Application: Active Learning for Robot Pushing}
\label{sec:real-world-robots}

\begin{figure}[t]
  \centering
  \begin{minipage}[t]{0.3\textwidth}
    \centering
    \includegraphics[width=\textwidth, clip, trim={28 28 28 28}]{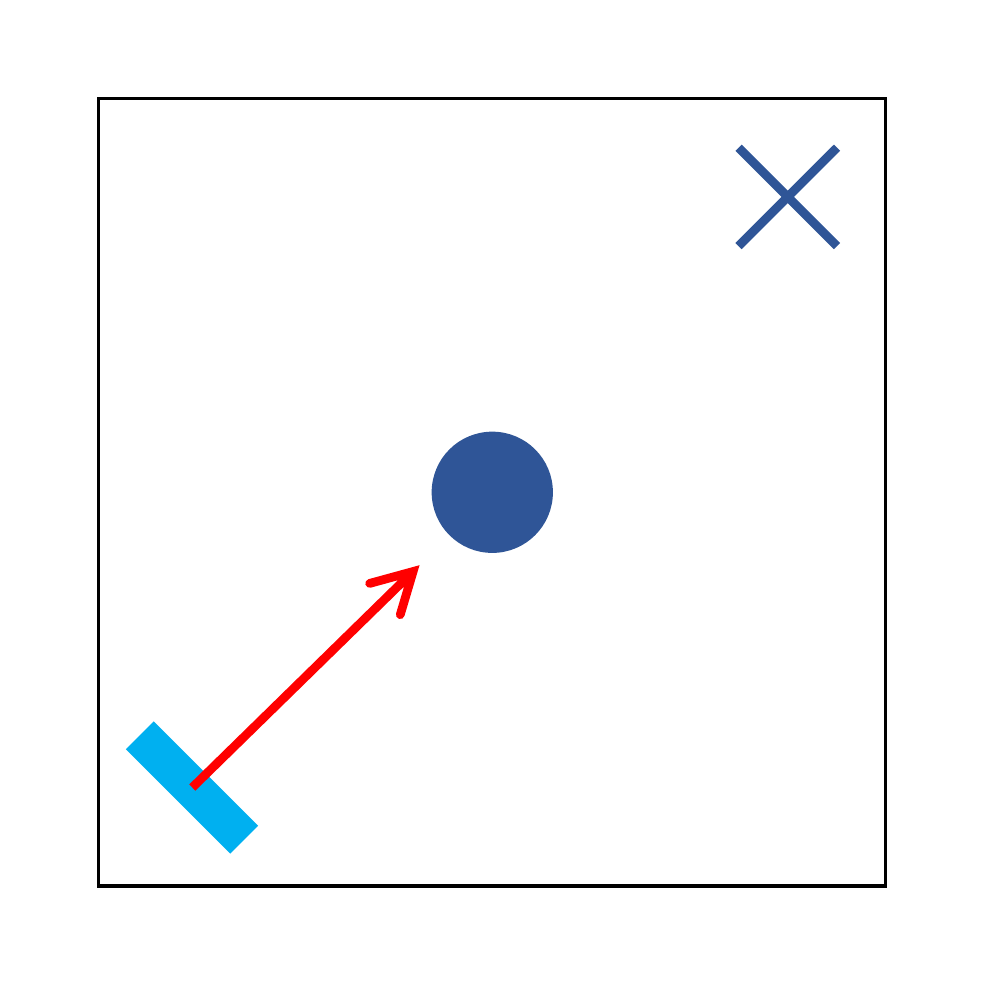}
    \textsc{push4}
  \end{minipage}%
  \hfil%
  \begin{minipage}[t]{0.3\textwidth}
    \centering
    \includegraphics[width=\textwidth, clip, trim={28 28 28 28}]{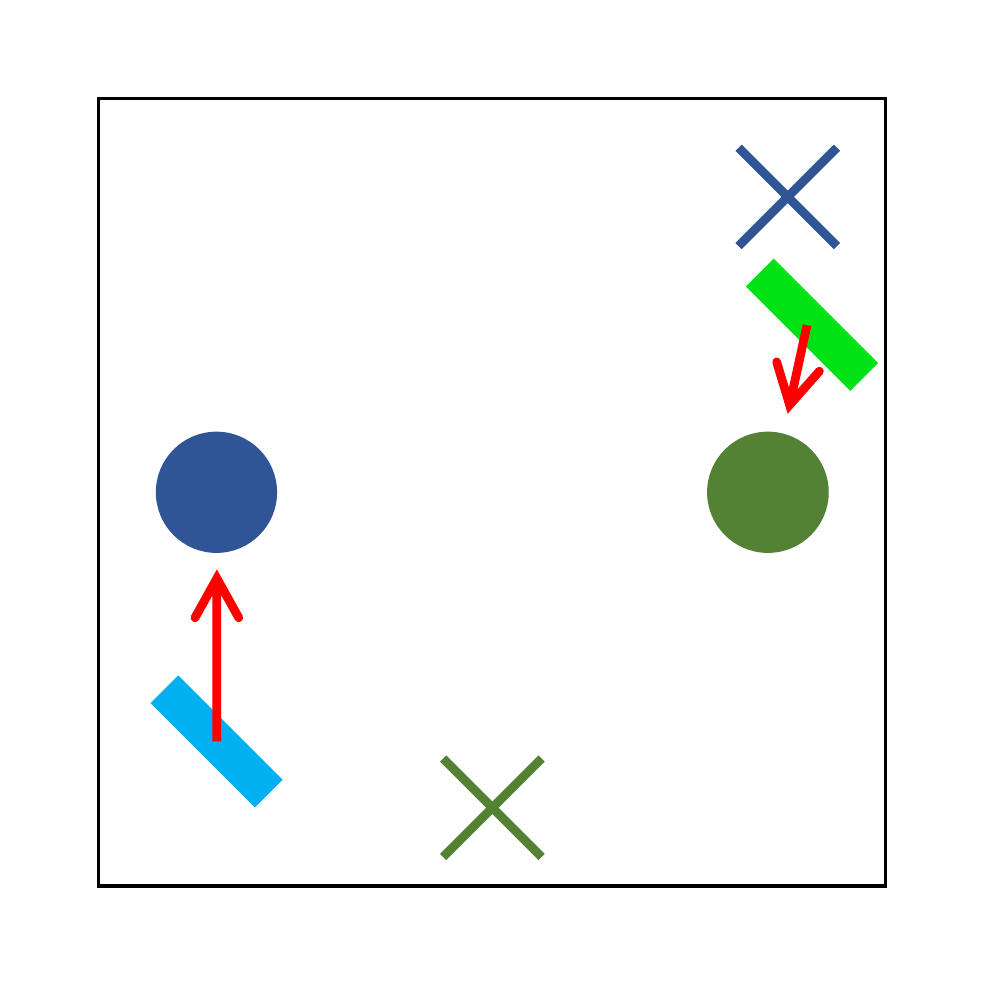}
    \textsc{push8}
  \end{minipage}%
  \caption{Two robot pushing tasks. \textsc{push4} (\textit{left}): a robot
    hand (rectangle) pushes the object (circle) towards a target (cross) in an
    unknown location. As indicated by the arrows, the
    robot always travels in the direction of the
    object's initial location and only receives feedback in the form of the
    distance of the object, after pushing, to the target. \textsc{push8}
    (\textit{right}): Similarly, two robots push their objects towards
    unknown target locations. Note that in \textsc{push4} the robot is
    likely to push the ball close to the target because it is initially
    positioned well and has its hand orientated towards the object. In
    contrast, neither robot in \textsc{push8} is likely to push its object
    close to the target because each begins in a worse location and is not
    orientated in a manner conducive to pushing. }
\label{fig:robot_tasks}
\end{figure}

Following \citet{wang:MES} and \citet{jiang:nonmyopicbo}, we optimise the
control parameters for two active learning robot pushing problems
\citep{wang:robots}. In the first problem, illustrated in
Figure~\ref{fig:robot_tasks}, a robot hand (rectangle) is given the task of
pushing an object (circle) towards an unknown target location (cross). Once
the robot has pushed the object it receives feedback in the form the
distance of the object to the target. The robot's movement is constrained
such that it can only travel in the direction of the object's initial
location. Adjustable parameters are the robot's starting position, the
orientation of its hand and the length of time it travels. This can
therefore be viewed as minimisation problem in which these four parameters
are optimised to minimise the distance of the object's final location to
the target. We denote the resulting four-dimensional problem
\textsc{push4}.

In the second problem, \textsc{push8}, shown in
Figure~\ref{fig:robot_tasks}, two robots (blue and green rectangles) in the
same arena have to push their respective objects (circles) towards unknown
targets (crosses). Their movements are constrained similarly to
\textsc{push4}, meaning that if they are initialised facing one another
they will block each other's path. The final distances of each of the
pushed objects to the corresponding target are summed and the total is used as the
feedback for both robots, resulting in a joint learning task. We treat this
as a minimisation problem: the 8 parameters determining the robots' paths
are to be optimised to minimise the combined distance of the objects to
their targets.

Like \citet{wang:MES}, the object's initial location in \textsc{push4} is
always the centre of the domain and the target location is changed on each
optimisation run. Corresponding runs for each optimisation method used the
same target location so that the runs were directly comparable. The targets'
positions were selected by Latin hypercube sampling of 51 positions across
the domain. We thus average over instances of the problem class, rather
than repeatedly optimising the same function from different
initialisations --- this supports the assessment of results generalised to 
starting positions  (see \citep{bartz:create}  for a broader discussion on 
problem generators and generalisable results). Likewise, in \textsc{push8} the
object's initial locations were fixed as shown in Figure~\ref{fig:robot_tasks}
and each target's positions were generated in the same way as the
\textsc{push4} targets. Target positions were paired such that the minimum
distance between the targets for each problem instance was sufficient for
the objects to be placed on the targets without overlapping. However, this
does not mean that in each instance it is possible for the robots to
actually push the objects to their targets because the targets may be
positioned so that the robots would block each other \textit{en route} to
their targets.  Since this means that the optimum distance for some of
these problem instances is not zero, in order to report the difference
between the optimised function value and the optimum we sought the global
optimum of each problem instance by randomly sampling the feasible space with 
$10^5$ sets of robot parameters and locally optimised the 100 best of these 
with the L-BFGS-B algorithm \citep{byrd:lbfgs}.  In fact, several of the
optimisation runs discovered better solutions than this procedure and in
these cases we used the resulting value as the estimate of the global
optimum. 

\begin{figure}[t]
\centering
\includegraphics[width=\textwidth]{figs/convergence_LEGEND}\\
\includegraphics[width=\textwidth, clip, trim={7 11 0 7}]{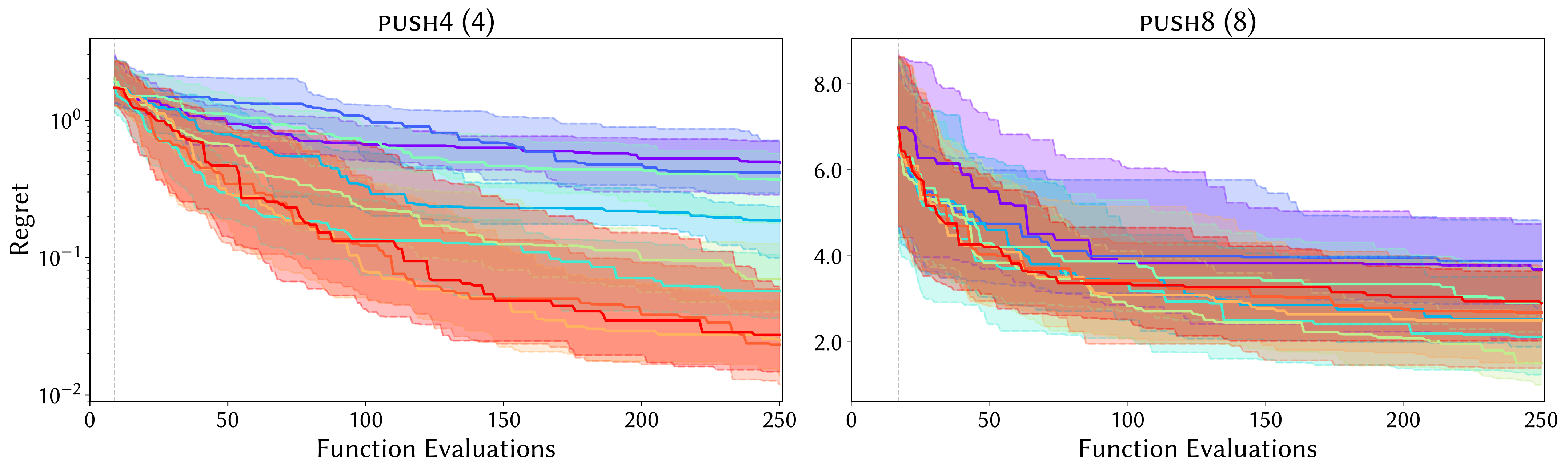}\\
\includegraphics[width=\textwidth, clip, trim={7 8 8 0}]{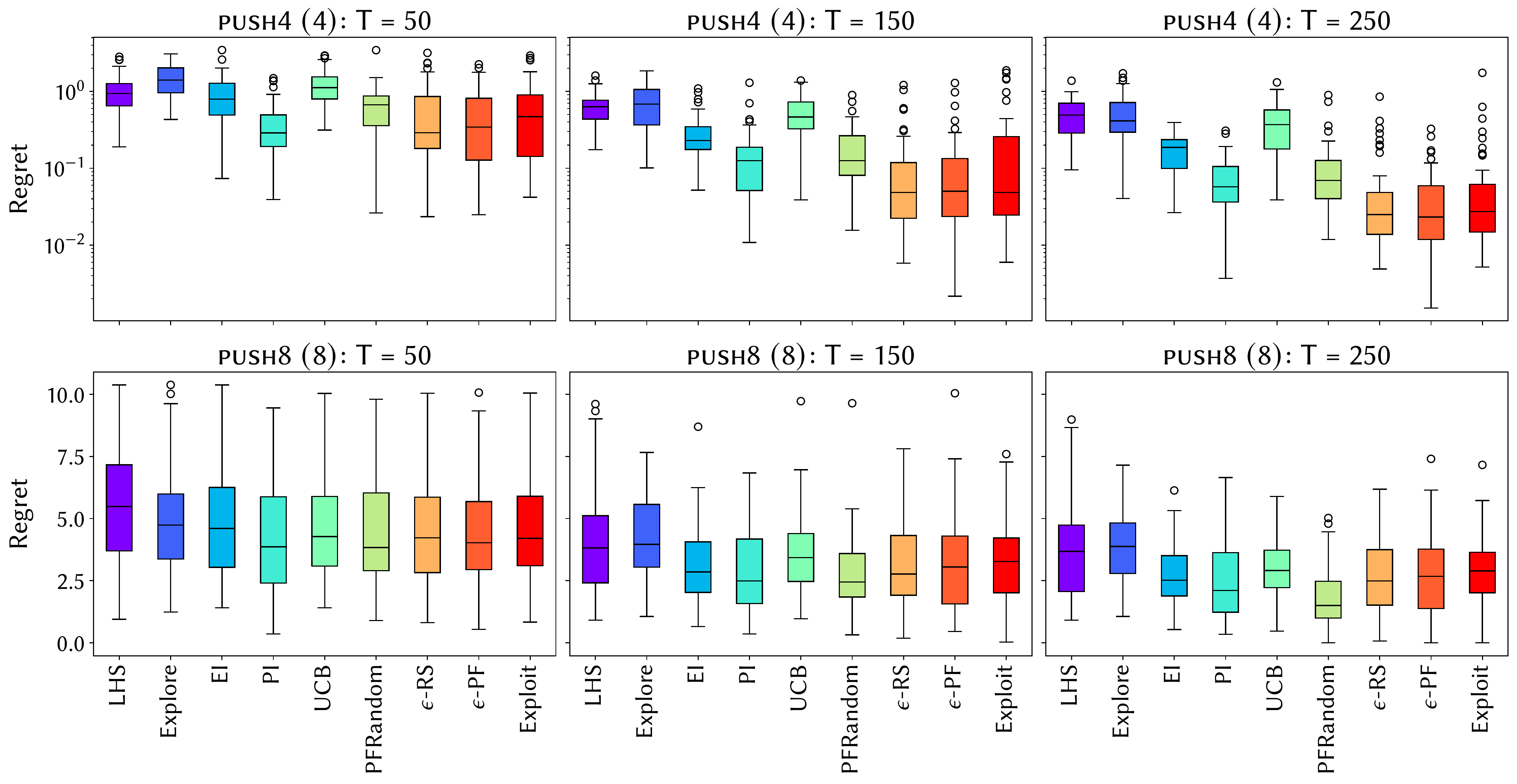}%
\caption{Illustrative convergence plots for the two robot pushing problems
(\textit{upper}) and the distribution of the best-seen function values (\textit{lower}) after 50 (\textit{left}),
150 (\textit{centre}), and 250 (\textit{right}) evaluations for both problems.}
\label{fig:convergence_push}
\end{figure}

Figure~\ref{fig:convergence_push} shows convergence histories and box plots
summarising the performance of each of the tested methods after 50, 150 and
250 function evaluations. 
As these results show, in the
four-dimensional \textsc{push4} problem, the exploitative methods
outperform the EI, PI and UCB acquisition functions. The \eFront method has
the median approach to the optimum, but \eRandom and pure exploitation are
statistically indistinguishable. In the harder \textsc{push8} problem all
of the optimisers are still far from the optimum, even after 250 function
evaluations.
Only random selection from the Pareto front (PFRandom) is significantly
better than any other method, and we note that PFRandom also performed well
in the 10-dimensional PitzDaily optimisation. We speculate that the
PFRandom, which selects from the entire Pareto front at each iteration,
owes its good performance to the additional exploration resulting from this
strategy, allowing it to explore the complicated optimisation landscape.
The \textsc{push8} optimisation landscape is particularly rugged and
difficult to approximate with Gaussian processes due to the abrupt changes
in fitness occurring as the robots' paths intersect. However, we note that
increasing exploration by increasing $\epsilon$ for the \eFront and
\eRandom methods does not significantly improve their performance. See the
supplementary material for these results as well as for videos of the best
solutions found to several of the problem instances evaluated.

\section{Conclusion}
\label{sec:conclusion}

How the balance between exploration and exploitation is chosen is clearer
in Bayesian optimisation than in some stochastic optimisation algorithms.
We have shown that the Expected Improvement and Upper Confidence Bound
acquisition functions select solutions from the Pareto optimal trade-off
between exploration and exploitation. However, the both the Weighted
Expected Improvement (for $\omega$ not in the range $ (0.185, 0.5]$) and Probability of Improvement
function may choose dominated solutions.  This  may account for the poor
empirical performance of the PI acquisition function.

Our analysis and experiments indicate that an effective strategy
is to be mostly greedy, occasionally selecting a random exploratory
solution. $\epsilon$-greedy acquisition functions that select from either
the Pareto front of maximally exploratory and exploitative solutions or the
entire feasible space perform almost equivalently and the algorithms are not
sensitive to the precise value of $\epsilon$. The need for exploration via
deliberate inclusion of exploratory moves turns out to be less important as
the dimension of decision space increases and the purely exploitative
method is fortuitously exploratory because of the low fidelity surrogate
modelling; improving the quality of surrogate models in the face of the
curse of dimensionality is an important topic of future research. While
$\epsilon$-greedy algorithms are trivially guaranteed to converge
eventually, we look forward to theoretical results on the rate of
convergence.

\begin{acks}
\label{sec:acknowledgement}
We thank Dr Steven Daniels for helping us prepare
Figure~\ref{fig:streamlines_pitzdaily}. 
This work was supported by Innovate UK grant number 104400.
\end{acks}
\bibliographystyle{ACM-Reference-Format}
\bibliography{ref}
\end{document}


\title{Supplementary Material for Greed is Good: Exploration and Exploitation Trade-offs in Bayesian Optimisation}

\author{George {De Ath}}
\email{g.de.ath@exeter.ac.uk}
\orcid{0000-0003-4909-0257}
\affiliation{%
  \department{Department of Computer Science}
  \institution{University of Exeter}
  \city{Exeter}
  \country{United Kingdom}
}

\author{Richard M. Everson}
\email{r.m.everson@exeter.ac.uk}
\orcid{0000-0002-3964-1150}
\affiliation{%
  \department{Department of Computer Science}
  \institution{University of Exeter}
  \city{Exeter}
  \country{United Kingdom}
}

\author{Alma A. M. Rahat}
\email{a.a.m.rahat@swansea.ac.uk}
\orcid{0000-0002-5023-1371}
\affiliation{%
  \department{Department of Computer Science}
  \institution{Swansea University}
  \city{Swansea}
  \country{United Kingdom}
}

\author{Jonathan E. Fieldsend}
\email{j.e.fieldsend@exeter.ac.uk}
\orcid{0000-0002-0683-2583}
\affiliation{%
  \department{Department of Computer Science}
  \institution{University of Exeter}
  \city{Exeter}
  \country{United Kingdom}
}

\maketitle

\appendix

\section{Synthetic function details}
\label{sec:synth_funcs}
In the following section we give the formulae of each of the 10 synthetic
functions optimised in this work. Where functions have been modified from their
standard form, we label the original functions as $g(\bx)$ and minimised
function as $f(\bx)$. In the cases where the functions are logged and their
minimum value can be negative, we add a constant value before the log
transformation to ensure that the minimum value of the function will always be
positive. Note that the value of the added constant does not affect the 
function's landscape.

\subsection{WangFreitas}
\begin{align}
g(x) &=  2 \exp \left( -\frac{1}{2} \left(\frac{x - a}{\theta_1}\right)^2 \right)
       + 4 \exp \left( -\frac{1}{2} \left(\frac{x - b}{\theta_2}\right)^2 \right)
\label{eqn:wangfreitas_original} \\
f(x) &= -g(x),
\label{eqn:wangfreitas}
\end{align}
where $a=0.1$, $b=0.9$, $\theta_1=0.1$ and $\theta_2=0.01$.

\subsection{Branin}
\begin{equation}
f(\bx) = a(x_2 - b x_1^2 + c x_1 - r)^2 + s(1 - t) \cos(x_1) + s,
\label{eqn:branin}
\end{equation}
where $a=1$, $b=\tfrac{5.1}{4 \pi^2}$, $c=\tfrac{5}{\pi}$, $r=6$, $s=10$, 
$t=\tfrac{1}{8 \pi}$ and $x_i$ refers to the $i$-th element of $\bx$.

\subsection{BraninForrester}
\begin{equation}
f(\bx) = a(x_2 - b x_1^2 + c x_1 - r)^2 + s(1 - t) \cos(x_1) + s + 5 x_1,
\label{eqn:braninforrester}
\end{equation}
where $a=1$, $b=\tfrac{5.1}{4 \pi^2}$, $c=\tfrac{5}{\pi}$, $r=6$, $s=10$, 
and $t=\tfrac{1}{8 \pi}$.

\subsection{Cosines}
\begin{align}
g(\bx) =& ~ 1 - \sum_{i=1}^2 \left[ (1.6 x_i - 0.5)^2 - 0.3 \cos (3 \pi (1.6 x_i - 0.5)) \right] \\
f(\bx) =& - g(\bx).
\label{eqn:cosines}
\end{align}

\subsection{logGoldsteinPrice}
\begin{align}
\begin{split}
g(\bx) = ~ & (1 + (x_1 + x_2 +1 )^2 (19 - 14 x_1 + 3 x_1^2 - 14 x_2 + 6 x_1 x_2 + 3 x_2^2)) \\
         & \times (30 + (2 x_1 - 3 x_2)^2 (18 - 32 x_1 + 12 x_1^2 + 48 x_2 - 36 x_1 x_2 + 27 x_2^2))
\end{split} \\
f(\bx) = & \log \left( g(\bx)\right).
\label{eqn:loggoldsteinprice}
\end{align}

\subsection{logSixHumpCamel}
\begin{align}
g(\bx) = & (4 - 2.1 x_1^2 + \frac{x_1^4}{3}) x_1^2 + x_1 x_2 + (-4 + 4 x_2^2) x_2^2 \\
f(\bx) = & \log \left( g(\bx)  + a + b \right),
\label{eqn:logsixhumpcamel}
\end{align}
where $a = 1.0316$ and $b = 10^{-4}$. $g(\bx)$ has a minimum value of 
$-1.0316$ and, therefore, we add $a$ plus a small constant $b$.

\subsection{logHartmann6}
\begin{align}
g(\bx) = & -\sum_{i=1}^4 \alpha_i \exp \left( -\sum_{j=1}^6 A_{i j} \left(x_j - P_{i j} \right)^2 \right) \\
f(\bx) = & - \log \left( - g(\bx) \right)
\label{eqn:loghartmann6}
\end{align}
where
\begin{align}
\mathbf{\alpha} = & \left( 1.0, 1.2, 3.0, 3.2 \right)^T \\
\mathbf{A} = & \left(\begin{array}{cccccc} 10   & 3   & 17   & 3.50 & 1.7 & 8 \\ 
                                           0.05 & 10  & 17   & 0.1  & 8   & 14 \\ 
                                           3    & 3.5 & 1.7  & 10   & 17  & 8 \\ 
                                           17   & 8   & 0.05 & 10   & 0.1 & 14
                     \end{array}\right) \\
\mathbf{P} = & 10^{-4} \left(\begin{array}{cccccc} 1312 & 1696 & 5569 &  124 & 8283 & 5886 \\ 
                                                   2329 & 4135 & 8307 & 3736 & 1004 & 9991 \\ 
                                                   2348 & 1451 & 3522 & 2883 & 3047 & 6650 \\ 
                                                   4047 & 8828 & 8732 & 5743 & 1091 &  381 
                             \end{array}\right).
\label{eqn:loghartmann6_variables}
\end{align}

\subsection{logGSobol}
\begin{align}
g(\bx) = & \prod_{i=1}^D \frac{4 x_i - a_i}{2} \\
f(\bx) = & \log \left( g(\bx) \right),
\label{eqn:loggsobol}
\end{align}
where $a_i = 1 \, \forall i \in \{1, 2, \dots, D\}$ and $D = 10$.

\subsection{logRosenbrock}
\begin{align}
g(\bx) = & \sum_{i=1}^{D-1} \left[ 100 (x_{i+1} - x_i^2)^2 + (x_i - 1)^2 \right] \\
f(\bx) = & \log \left( g(\bx) + 0.5 \right),
\label{eqn:logrosenbrock}
\end{align}
where $D = 10$. Note, similarly to logSixHumpCamel, because $g(\bx)$ has a 
minimum value of $0$, we add a small value to ensure it is always positive.

\subsection{logStyblinskiTang}
\begin{align}
g(\bx) = & \frac{1}{2} \sum_{i=1}^D (x_i^4 - 16 x_i^2 + 5 x_i) \\
f(\bx) = & \log \left( g(\bx) + 40D \right),
\label{eqn:logstyblinskitang}
\end{align}
where $D = 10$. Because $g(\bx)$ has a minimum value of $-39.16599D$, we add
$40D$ to it to ensure it is always positive.

\section{The Landscape of the WangFreitas Test Problem}
\begin{figure}[H]
\centering
\includegraphics[width=\textwidth]{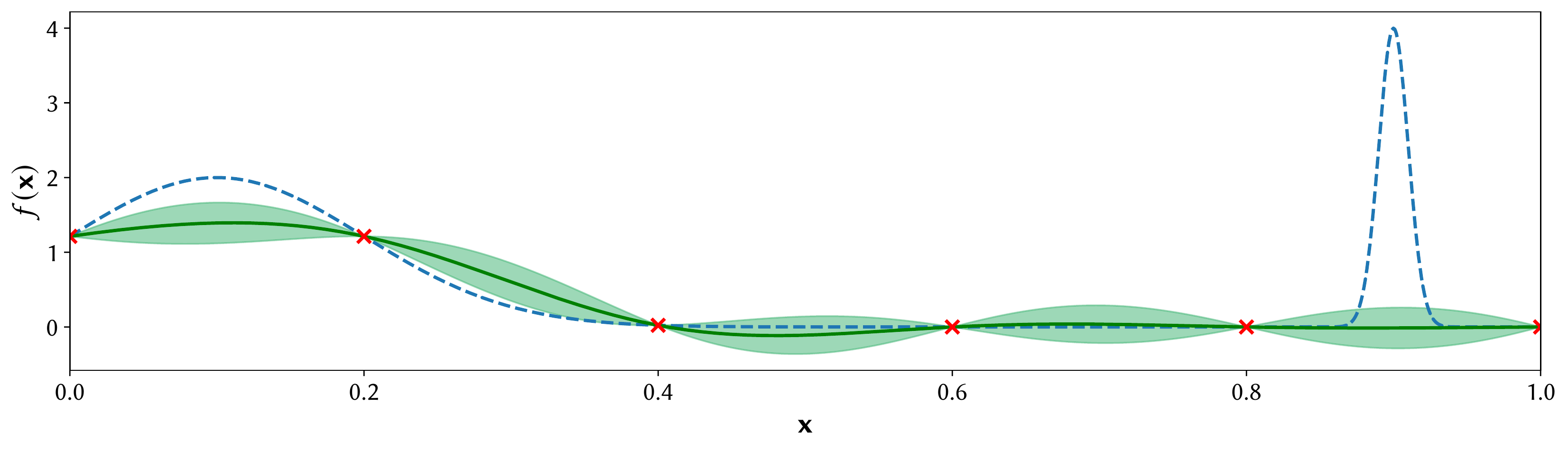}
\caption{The WangFreitas test problem. The blue line shows the true function, 
the green solid line shows the mean prediction of a \mGP model trained on the 
red crosses, and the green areas depict the uncertainty (twice the standard 
deviation).}
\label{fig:WangFreitasFunction}
\end{figure}
Figure~\ref{fig:WangFreitasFunction} shows an illustration of the test problem
(Equation~\ref{eqn:wangfreitas_original}) proposed by \citet{wangfreitas:2014}.
It has one local optimum and a global optimum. The global optimum has a narrow
basin surrounded by vast flat regions. Therefore it is easy for the model to
become overconfident about the flatness in the vicinity of the optimum with no
data identifying the basin, and mislead the search away from it. Consequently,
methods with high exploration do well in solving this problem.

\section{Full Experimental Results}
In this section we show the results table for the PitzDaily and robot pushing
problems, and the convergence and box plots for all test problems evaluated in
this work.

\subsection{PitzDaily Results Table}
  \begin{table}[H]
  \centering
  \setlength{\tabcolsep}{2pt}
  \sisetup{table-format=1.2e-1,table-number-alignment=center}
  \begin{tabular}{l SS}
    \toprule
    \bfseries Method
    & \multicolumn{2}{c}{\bfseries PitzDaily (10)} \\ 
    & \multicolumn{1}{c}{Median} & \multicolumn{1}{c}{MAD}  \\ \midrule
    Uniform & 9.58e-02 & 3.52e-03 \\
    Explore & 8.82e-02 & 4.82e-03 \\
    EI & \statsimilar 8.42e-02 & \statsimilar 1.43e-03 \\
    PI & 9.66e-02 & 4.96e-03 \\
    UCB & 8.55e-02 & 2.96e-03 \\
    PFRandom & \best 8.36e-02 & \best 9.72e-04 \\
    \eRandom & \statsimilar 8.49e-02 & \statsimilar 2.68e-03 \\
    \eFront & \statsimilar 8.45e-02 & \statsimilar 2.44e-03 \\
    Exploit & \statsimilar 8.40e-02 & \statsimilar 1.82e-03 \\
\bottomrule
  \end{tabular}
   \caption{Median absolute distance (\textit{left}) and median absolute deviation from
	the median (MAD, \textit{right}) from the optimum after 250 function evaluations, 
	across the 51 runs. The method with the lowest median performance is 
	shown in dark grey, with those with statistically equivalent 
	performance are shown in light grey.}
  \label{tbl:synthetic_results-pitzdaily}
  \end{table}
The full results of the optimisation runs on the PitzDaily test problem are 
shown in Table~\ref{tbl:synthetic_results-pitzdaily}. It shows the median 
difference between the estimated optimum and the true optimum over the 51 
repeated experiments, together with the median absolute deviation from the 
median (MAD). The method with the minimum median on each function is 
highlighted in dark grey, and those which are statistically equivalent to the best 
method according to a one-sided paired Wilcoxon signed-rank test 
\citep{knowles:testing} with Holm-Bonferroni correction \citep{holm:test}
($p\geq0.05$), are shown in light grey.

\subsection{Robot Pushing Results Table}
  \begin{table}[H]
  \centering
  \setlength{\tabcolsep}{2pt}
  \begin{tabular}{l S[table-format=1.2e-1]S[table-format=1.2e-1] 
                    S[table-format=1.2e0]S[table-format=1.2e0]}
    \toprule
    \bfseries Method
    & \multicolumn{2}{c}{\bfseries \textsc{push4} (4)} 
    & \multicolumn{2}{c}{\bfseries \textsc{push8} (8)} \\ 
    & \multicolumn{1}{c}{Median} & \multicolumn{1}{c}{MAD}
    & \multicolumn{1}{c}{Median} & \multicolumn{1}{c}{MAD}  \\ \midrule
    LHS & 4.93e-01 & 3.08e-01 & 3.68e+00 & 2.18e+00 \\
    Explore & 4.14e-01 & 2.41e-01 & 3.88e+00 & 1.44e+00 \\
    EI & 1.86e-01 & 1.05e-01 & 2.52e+00 & 1.07e+00 \\
    PI & 5.72e-02 & 4.45e-02 & 2.11e+00 & 1.47e+00 \\
    UCB & 3.70e-01 & 2.90e-01 & 2.91e+00 & 1.19e+00 \\
    PFRandom & 6.95e-02 & 6.71e-02 & \best 1.50e+00 & \best 1.07e+00 \\
    \eRandom & \statsimilar 2.50e-02 & \statsimilar 2.17e-02 & 2.49e+00 & 1.56e+00 \\
    \eFront & \best 2.32e-02 & \best 2.47e-02 & 2.68e+00 & 1.80e+00 \\
    Exploit & \statsimilar 2.73e-02 & \statsimilar 2.51e-02 & 2.89e+00 & 1.23e+00 \\
\bottomrule
  \end{tabular}
  \caption{Median absolute distance (\textit{left}) and median absolute deviation from the
    median (MAD, \textit{right}) from the optimum after 250 function evaluations across 
    the 51 runs. The method with the lowest median performance is shown
    in dark grey, with those with statistically equivalent performance
    are shown in light grey. }
\label{tbl:push_results}
  \end{table}

The full results of the optimisation runs on the \textsc{push4} and 
\textsc{push8} test problems are 
shown in Table~\ref{tbl:push_results}.  It shows the median
difference between the estimated optimum and the true optimum over the 51 
repeated experiments, together with the median absolute deviation from the 
median (MAD). The method with the minimum median on each function is 
highlighted in dark grey, and those which are statistically equivalent to the best 
method according to a one-sided paired Wilcoxon signed-rank test 
\citep{knowles:testing} with Holm-Bonferroni correction \citep{holm:test}
($p\geq0.05$), are shown in light grey.

\subsection{Convergence Histories and Boxplots}

In this section we display the full set of results for the experimental
evaluations carried out in this paper. Each figure shows the convergence of 
each algorithm on the respective test problem (\textit{top}), snapshots of their 
performance at 50, 150, and 250 function evaluations (\textit{centre}), and the 
comparative performance between \eFront (green) and \eRandom (red, hatched) for
increasing values of $\epsilon$ (\textit{lower}). 

\newpage

\readlist*\problems{
WangFreitas, 
Branin, 
Cosines, 
BraninForrester, 
logGoldsteinPrice,
logSixHumpCamel, 
logHartmann6, 
logGSobol, 
logRosenbrock, 
logStyblinskiTang,
PitzDaily,
push4,
push8
}

\readlist*\legendnames{
LEGEND, 
LEGEND, 
LEGEND, 
LEGEND, 
LEGEND,
LEGEND, 
LEGEND, 
LEGEND, 
LEGEND, 
LEGEND,
LEGEND_PitzDaily,
LEGEND,
LEGEND
}

\readlist*\dims{
one-dimensional, 
two-dimensional, 
two-dimensional, 
two-dimensional, 
two-dimensional, 
two-dimensional, 
six-dimensional, 
ten-dimensional, 
ten-dimensional, 
ten-dimensional, 
ten-dimensional real-world,
four-dimensional real-world,
eight-dimensional real-world
}

\foreach \i in {1,...,\problemslen} {
	\def \problem {\problems[\i]}
	\def \dim {\dims[\i]\xspace}
	\def \legendname {\legendnames[\i]}
	\def \captn {
		Results for the \dim \problem ~test problem. The convergence histories for 
		each algorithm are shown in the upper figure, where the shaded regions 
		correspond to the interquartile range. The central figure shows the 
        distribution of best seen function evaluations after 50 
        (\textit{left}), 150 (\textit{centre}) and 250 (\textit{right}) 
        function evaluations have occurred. The lower figure shows a
		comparison between \textcolor{OliveGreen}{\eFront} 
		(\textcolor{OliveGreen}{green}) and \textcolor{Maroon}{\eRandom} 
		(\textcolor{Maroon}{red, hatched}) for different values of $\epsilon$ 
        (horizontal axis) after 50 (\textit{left}), 150 (\textit{centre}) 
        and 250 (\textit{right}) function evaluations.
	}
	
	\begin{figure}[H]
	\centering
	\includegraphics[width=0.9\textwidth]{figs/convergence_\legendname}
	\includegraphics[width=0.9\textwidth]{figs/convergence_\problem}
	\includegraphics[width=\textwidth]{figs/boxplots_\problem}
	\includegraphics[width=\textwidth]{figs/egreedy_compare_\problem}
	\caption{\captn}
	\label{fig:\problem}
	\end{figure}
	\newpage
}

\section{Optimisation of the Synthetic functions without log transformation}
In order to illustrate the effect of improving the surrogate model by
log-transforming the functions with large scale changes in their observed
values, we also present results on the synthetic functions without the log
transformation. The functional form of each of the test functions are the same
as the values of $g(\bx)$ for each of their log-transformed counterparts in
Section~\ref{sec:synth_funcs}.

\subsection{Results Table}
The full results of the optimisation runs on the six test problems are shown in
Table~\ref{tbl:synthetic_results}. It shows the median difference between the
estimated optimum and the true optimum over the 51 repeated experiments,
together with the median absolute deviation from the median (MAD). The method
with the minimum median on each function is highlighted in dark grey, and those
which are statistically equivalent to the best method according to a one-sided
paired Wilcoxon signed-rank test \citep{knowles:testing} with Holm-Bonferroni
correction \citep{holm:test} ($p\geq0.05$), are shown in light grey.
\begin{table}[H]
    \setlength{\tabcolsep}{2pt}
    \sisetup{table-format=1.2e-1,table-number-alignment=center}
    \resizebox{0.85\textwidth}{!}{%
    \begin{tabular}{l SS SS SS}
    \toprule
    \bfseries Method
    & \multicolumn{2}{c}{\bfseries GoldsteinPrice (1)} 
    & \multicolumn{2}{c}{\bfseries SixHumpCamel (2)} 
    & \multicolumn{2}{c}{\bfseries Hartmann6 (2)} \\ 
    & \multicolumn{1}{c}{Median} & \multicolumn{1}{c}{MAD}
    & \multicolumn{1}{c}{Median} & \multicolumn{1}{c}{MAD}
    & \multicolumn{1}{c}{Median} & \multicolumn{1}{c}{MAD}  \\ \midrule
    LHS & 5.85e+00 & 6.39e+00 & 4.85e-02 & 5.32e-02 & 9.50e-01 & 2.71e-01 \\
    Explore & 4.13e+00 & 3.25e+00 & 5.75e-02 & 4.75e-02 & 8.51e-01 & 1.77e-01 \\
    EI & 1.82e-04 & 2.27e-04 & \best 9.57e-07 & \best 9.79e-07 & 1.30e-03 & 6.67e-04 \\
    PI & 3.30e-04 & 3.97e-04 & 2.02e-06 & 2.93e-06 & \statsimilar 1.06e-03 & \statsimilar 1.43e-03 \\
    UCB & 5.80e-02 & 8.50e-02 & 1.92e-06 & 2.42e-06 & 4.25e-01 & 1.26e-01 \\
    PFRandom & 5.79e-01 & 8.49e-01 & 2.63e-04 & 3.22e-04 & \statsimilar 2.93e-02 & \statsimilar 2.70e-02 \\
    \eRandom & \best 4.04e-05 & \best 5.44e-05 & \statsimilar 9.91e-07 & \statsimilar 9.20e-07 & \best 8.54e-04 & \best 6.31e-04 \\
    \eFront & \statsimilar 6.01e-05 & \statsimilar 8.30e-05 & \statsimilar 1.03e-06 & \statsimilar 1.10e-06 & \statsimilar 1.17e-03 & \statsimilar 6.63e-04 \\
    Exploit & 7.26e-05 & 1.07e-04 & \statsimilar 1.29e-06 & \statsimilar 1.59e-06 & \statsimilar 9.78e-04 & \statsimilar 9.51e-04 \\
\bottomrule
    \toprule
    \bfseries Method
    & \multicolumn{2}{c}{\bfseries GSobol (2)} 
    & \multicolumn{2}{c}{\bfseries Rosenbrock (6)} 
    & \multicolumn{2}{c}{\bfseries StyblinskiTang (10)} \\ 
    & \multicolumn{1}{c}{Median} & \multicolumn{1}{c}{MAD}
    & \multicolumn{1}{c}{Median} & \multicolumn{1}{c}{MAD}
    & \multicolumn{1}{c}{Median} & \multicolumn{1}{c}{MAD}  \\ \midrule
    LHS & 3.43e+03 & 3.24e+03 & 5.35e+04 & 2.75e+04 & 1.35e+02 & 2.65e+01 \\
    Explore & 3.81e+04 & 3.84e+04 & 1.76e+05 & 8.92e+04 & 1.94e+02 & 2.22e+01 \\
    EI & 4.68e+02 & 5.85e+02 & \statsimilar 1.35e+03 & \statsimilar 6.21e+02 & \statsimilar 4.48e+01 & \statsimilar 2.07e+01 \\
    PI & \statsimilar 3.45e+02 & \statsimilar 3.60e+02 & 2.26e+03 & 9.79e+02 & \best 4.24e+01 & \best 2.09e+01 \\
    UCB & 4.57e+02 & 4.49e+02 & 1.49e+03 & 7.34e+02 & 1.58e+02 & 2.20e+01 \\
    PFRandom & 9.18e+02 & 9.15e+02 & 4.45e+03 & 1.83e+03 & 1.21e+02 & 4.04e+01 \\
    \eRandom & \best 1.63e+02 & \best 1.94e+02 & \best 1.11e+03 & \best 5.89e+02 & \statsimilar 4.44e+01 & \statsimilar 2.18e+01 \\
    \eFront & 4.08e+02 & 5.08e+02 & \statsimilar 1.35e+03 & \statsimilar 7.04e+02 & \statsimilar 4.42e+01 & \statsimilar 1.89e+01 \\
    Exploit & 5.21e+02 & 6.46e+02 & \statsimilar 1.38e+03 & \statsimilar 6.61e+02 & 4.48e+01 & 2.25e+01 \\
\bottomrule
    \end{tabular}
    }
    \vspace*{0.1mm}
    \caption{Median absolute distance (\textit{left}) and median absolute deviation from the
    median (MAD, \textit{right}) from the optimum after 250 function evaluations across 
    the 51 runs. The method with the lowest median performance is shown
    in dark grey, with those with statistically equivalent performance
    are shown in light grey. }
    \label{tbl:synthetic_results}
    \end{table}  

\newpage

\subsection{Convergence Histories and Boxplots}
In this section we display the full set of results for the experimental
evaluations carried out. Each figure
shows the convergence of each algorithm on the respective test problem 
(\textit{top}), snapshots of their performance at 50, 150, and 250 function evaluations 
(\textit{centre}), and the comparative performance between \eFront (green) and \eRandom
(red, hatched) for increasing values of $\epsilon$ (\textit{lower}).

\newpage

\readlist*\problems{
GoldsteinPrice,
SixHumpCamel, 
Hartmann6, 
GSobol, 
Rosenbrock, 
StyblinskiTang
}

\readlist*\legendnames{
LEGEND, 
LEGEND, 
LEGEND, 
LEGEND, 
LEGEND,
LEGEND
}

\readlist*\dims{
two-dimensional, 
two-dimensional, 
six-dimensional, 
ten-dimensional, 
ten-dimensional, 
ten-dimensional
}

\foreach \i in {1,...,\problemslen} {
	\def \problem {\problems[\i]}
	\def \dim {\dims[\i]\xspace}
	\def \legendname {\legendnames[\i]}
	\def \captn {
		Results for the \dim \problem ~test problem. The convergence histories for 
		each algorithm are shown in the upper figure, where the shaded regions 
		correspond to the interquartile range. The central figure shows the 
        distribution of best seen function evaluations after 50 (\textit{left}), 
        150 (\textit{centre}) and 250 (\textit{right}) function evaluations 
        have occurred. The lower figure shows a 
		comparison between \textcolor{OliveGreen}{\eFront} 
		(\textcolor{OliveGreen}{green}) and \textcolor{Maroon}{\eRandom} 
		(\textcolor{Maroon}{red, hatched}) for different values of $\epsilon$ 
        (horizontal axis) after 50 (\textit{left}), 150 (\textit{centre}) 
        and 250 (\textit{right}) function evaluations.
	}
	
	\begin{figure}[H]
	\centering
	\includegraphics[width=0.9\textwidth]{figs/convergence_\legendname}
	\includegraphics[width=0.9\textwidth]{figs/convergence_\problem}
	\includegraphics[width=\textwidth]{figs/boxplots_\problem}
	\includegraphics[width=\textwidth]{figs/egreedy_compare_\problem}
	\caption{\captn}
	\label{fig:\problem}
	\end{figure}
	\newpage
}

\bibliographystyle{ACM-Reference-Format}
\bibliography{ref}